\documentclass[journal]{IEEEtran}

\usepackage{xfrac}
\usepackage[T1]{fontenc}
\usepackage{verbatim}
\usepackage{float}
\usepackage{amsthm}
\usepackage{amsmath}
\usepackage{amssymb}
\usepackage{graphicx}
\usepackage{mathtools}
\usepackage[linesnumbered,ruled,vlined ]{algorithm2e}
\usepackage{color}
\usepackage{subfig}
\usepackage{url}
\usepackage{cite}

\newcommand{\vect}[1]{\boldsymbol{\mathbf{#1}}}

\newcommand{\diag}{\operatorname{diag}}

\newcommand{\R}{\mathbb{R}}
\newcommand{\ep}{\varepsilon}
\newcommand{\E}{\mathbb{E}}
\newcommand{\C}{\textnormal{Cov}}

\newcommand{\Po}{\textnormal{Poisson}}

\newcommand{\tr}{\textnormal{tr}}

\newcommand{\bitem}{\begin{itemize}}
\newcommand{\eitem}{\end{itemize}}
\newcommand{\benum}{\begin{enumerate}}
\newcommand{\eenum}{\end{enumerate}}
\newcommand{\beq}{\begin{equation}}
\newcommand{\eeq}{\end{equation}}
\newcommand{\beqs}{\begin{equation*}}
\newcommand{\eeqs}{\end{equation*}}

\begin{document}
\title{Steerable $e$PCA: Rotationally Invariant Exponential Family PCA}

\author{Zhizhen~Zhao~\IEEEmembership{Member,~IEEE,} Lydia~T.~Liu,
        and~Amit~Singer
\thanks{ZZ is with the Department
of Electrical and Computer Engineering, University of Illinois at Urbana-Champaign, Urbana, IL, 61820 USA e-mail: zhizhenz@illinois.edu .}
\thanks{LTL is with the Department of Electrical Engineering and Computer Sciences, University of California at Berkeley, Berkeley, CA, 94720 e-mail: lydiatliu@berkeley.edu}
\thanks{AS is with the Department of Mathematics, and Program in Applied and Computational Mathematics, Princeton University, Princeton, NJ, 08544 e-mail:amits@math.princeton.edu}
}


\maketitle

\begin{abstract}
In photon-limited imaging, the pixel intensities are affected by photon count noise. Many applications, such as 3-D reconstruction using correlation analysis in X-ray free electron laser (XFEL) single molecule imaging, require an accurate estimation of the covariance of the underlying 2-D clean images. Accurate estimation of the covariance from low-photon count images must take into account that pixel intensities are Poisson distributed, 
hence the classical sample covariance estimator is sub-optimal. Moreover, in single molecule imaging, including in-plane rotated copies of all images could further improve the accuracy of covariance estimation. In this paper we introduce an efficient and accurate algorithm for covariance matrix estimation of count noise 2-D images, including their uniform planar rotations and possibly reflections. Our procedure, {\em steerable $e$PCA}, combines in a novel way two recently introduced innovations. The first is a methodology for principal component analysis (PCA) for Poisson distributions, and more generally, exponential family distributions, called $e$PCA. The second is steerable PCA, a fast and accurate procedure for including all planar rotations for PCA. The resulting principal components are invariant to the rotation and reflection of the input images. We demonstrate the efficiency and accuracy of steerable $e$PCA in numerical experiments involving simulated XFEL datasets and rotated Yale B face data. 
\end{abstract}



\section{Introduction}
X-ray free electron laser (XFEL) is an emerging imaging technique for elucidating the three-dimensional structure of molecules~\cite{favre2015xtop, maia2016trickle}. Single molecule XFEL imaging collects two-dimensional diffraction patterns of single particles at random orientations.
The images are very noisy due to the low photon count and the detector count-noise follows an approximately Poisson distribution. Since only one diffraction pattern is captured per particle and the particle orientations are unknown, it is challenging to reconstruct the 3-D structure at low signal-to-noise ratio (SNR). One approach is to use expectation-maximization (EM) \cite{scheres2007disentangling, duane2009}, but it has a high computational cost at low signal-to-noise ratio (SNR). 
Alternatively, assuming that the orientations of the particles are uniformly distributed over the special orthogonal group $SO(3)$, Kam's correlation analysis~\cite{kam1977determination, Saldin2009,starodub2012single,pande2014deducing,
kurta2017correlations,donatelli2017reconstruction,vonArdenne2018,Pande11772} bypasses orientation estimation and requires just one pass over the data, thus alleviating the computational cost. Kam's method requires a robust estimation of the covariance matrix of the noiseless 2-D images. This serves as our main motivation for developing efficient and accurate covariance estimation and denoising methods for Poisson data. Nevertheless, the methods presented here are quite general and can be applied to other imaging modalities involving Poisson distributions.

Principal Component Analysis (PCA) is widely used for dimension reduction and denoising of large datasets~\cite{jolliffe2002principal,anderson1958introduction}. However, it is most naturally designed for Gaussian data, and there is no commonly agreed upon extension to non-Gaussian settings such as exponential family distributions~\cite[Sec. 14.4]{jolliffe2002principal}.  For denoising with non-Gaussian noise, popular approaches reduce it to the Gaussian case by a wavelet transform such as a Haar transform \cite{nowak1999wavelet}; by adaptive wavelet shrinkage~\cite{nowak1999wavelet, luisier2007new}; or by approximate variance stabilization such as the Anscombe transform~\cite{anscombe1948}. The latter is known to work well for Poisson signals with large parameters, due to approximate normality. However, the normal approximation breaks down for Poisson distributions with a small parameter, such as photon-limited XFEL \cite[Sec. 6.6]{starck2010sparse}. Other methods are based on alternating minimization \cite{Collins2001,singh08unified, salmon2014poisson}, singular value thresholding (SVT)~\cite{Furnival2016, Cao2014} and Bayesian techniques~\cite{Sonnleitner2016}.
Many of aforementioned methods such as \cite{Collins2001} are computationally intractable for large datasets and do not have statistical guarantees for covariance estimation. In particular, \cite{salmon2014poisson} applied the methodology of \cite{Collins2001} to denoise a single image by performing PCA on clusters of patches extracted from the image; it is not suitable for our problem setting, where the goal is to simultaneously denoise a large number of images and estimate their covariance. Recently, \cite{LiuEtAl2017} introduced exponential family PCA ($e$PCA), which extends PCA to a wider class of distributions. It involves the eigendecomposition of a new covariance matrix estimator, constructed in a deterministic and non-iterative way using moment calculations and shrinkage. $e$PCA was shown to be more accurate than PCA and its alternatives for exponential families. Its computational cost is similar to that of PCA, it has substantial theoretical justification building on random matrix theory, and its output is interpretable. We refer readers to \cite{LiuEtAl2017} for experiments that benchmark $e$PCA against previous methods.  

In XFEL imaging, the orientations of the particles are uniformly distributed over $SO(3)$, and it is therefore equally likely to observe 
any planar rotation of the given 2-D diffraction pattern. Therefore, it makes sense to include all possible in-plane rotations of the images when performing $e$PCA. To this end, we incorporate steerability in $e$PCA, by adapting the steerable PCA algorithm that avoids duplicating rotated images~\cite{zhao2016fast}. The concept of a steerable filter was introduced in~\cite{freeman1991design} and various methods were proposed for computing data adaptive steerable filters~\cite{Hilai,Perona,Uenohara,Ponce}. We take into account the action of the group $O(2)$ on diffraction patterns by in-plane rotation (and reflection if needed). The resulting principal components are invariant to any $O(2)$ transformation of the input images.

The new algorithm, to which we refer as steerable $e$PCA,  combines $e$PCA and steerable PCA in a natural way. Steerable $e$PCA does not involve any optimization, with all steps involve only basic linear algebra operations. The various steps include expansion in a steerable basis, eigen-decomposition, eigenvalue shrinkage, and different normalizations. The mathematical and statistical rationale for all steps is provided in Section~\ref{sec:methods}.

We illustrate the improvement in estimating the covariance matrix through image denoising in Section~\ref{sec:results}. 
Specifically, we introduce a Wiener-type filtering using the principal components.
Rotation invariance enhances the effectiveness of $e$PCA in covariance estimation and thus achieves better denoising. In addition, the denoised expansion coefficients are useful in building rotationally invariant image features (i.e. bispectrum-like features~\cite{Zhao14}). We verify the benefits of steerable $e$PCA through numerical experiments with simulated XFEL diffraction patterns and a natural image dataset--Yale B face database ~\cite{lee2005acquiring,georghiades2001few}. Similar to the case of standard PCA, the computational complexity of the steerable $e$PCA is lower than $e$PCA. 

An implementation of steerable $e$PCA in MATLAB is  publicly available at \url{github.com/zhizhenz/sepca/}. 

\section{Methods}
\label{sec:methods}
The goal of steerable $e$PCA is to estimate the rotationally invariant covariance matrix from the images whose pixel intensities are affected by photon count noise. To develop this estimator, we combine our previous works on steerable PCA with $e$PCA in a novel way. The main challenge in combining $e$PCA with steerable PCA is that steerable PCA involves a (nearly orthogonal) change of basis transformation from a Cartesian grid to a steerable basis. While multidimensional Gaussian distributions are invariant to orthogonal transformations, the multidimensional Poisson distribution is not invariant to such transformations. Therefore, steerable PCA and $e$PCA cannot be naively combined in a straightforward manner. Algorithm~\ref{sePCA_alg} details the steps of the combined procedure that achieves the goal of estimating a rotationally invariant covariance matrix from Cartesian grid images with pixel intensities following a Poisson distribution. The following subsections explain the main concepts underlying $e$PCA and steerable PCA, and how they are weaved together in a manner that overcomes the aforementioned challenge. In the following subsections, we detail the associated
concepts and steps for steerable $e$PCA.

\subsection{The observation model and homogenization}
\label{sec:obs_mod}
We adopt the same observation model introduced in~\cite{LiuEtAl2017}. We observe $n$ noisy images $Y_i\in \R^p$ (i.e., $p$ is the number of pixels), for $i = 1, \dots, n$. These are random vectors sampled from a hierarchical model defined as follows. First, a latent vector---or hyperparameter---$ \omega \in \R^p$ is drawn from a probability distribution $P$ with mean $\mu_{\omega}$ and covariance matrix $\Sigma_{\omega}$. Conditional on $\omega$, each coordinate of $Y = (Y(1),\ldots,Y(p))^\top$ is drawn independently from a canonical one-parameter exponential family, 
\beq
Y(j)|\omega(j)  \sim p_{\omega(j)}(y),  \quad Y = (Y(1),\ldots,Y(p))^\top,
\eeq
with density
\beq
\label{ef_dist}
p_{\omega(j)}(y)  = \exp[\omega(j) y - A( \omega (j))]
\eeq
with respect to a $\sigma$-finite measure $\nu$ on $\mathbb{R}$, where the $j^{\text{th}}$ entry of the latent vector, $\omega(j) \in \mathbb{R}$, is the natural parameter of the family and $A(\omega(j)) = \log \int
\exp(\omega(j) y) d\nu(y) $ is the corresponding log-partition function. 
The mean and variance of the random variable $Y(j)$ can be expressed as $A'(\omega(j))$ and $A''(\omega(j))$, where we denote $A'(\omega) = dA(\omega)/d \omega$. Therefore, the mean of $Y$ conditional on $\omega$ is
$$X := \E(Y|\omega) = (A'(\omega(1)),\ldots,A'(\omega(p)))^\top = A'(\omega),$$
so the noisy data vector $Y$ can be expressed as $Y= A'(\omega)+ \tilde\ep$, with $\E(\tilde \ep|\omega)=0$ and the marginal mean of $Y$ is $\E Y =\E A'(\omega)$. Thus, one can think of $Y$ as a noisy realization of the clean vector $X=A'(\omega)$. However, the latent vector $\omega$ is also random and varies from sample to sample. In the XFEL application, $X=A'(\omega)$ are the unobserved noiseless images, and their randomness stems for the random (and unobserved) orientation of the molecule. We may write\footnote{$\diag[x]$ for $x \in \R^p$ denotes a $p \times p$ diagonal matrix whose diagonal entries are $x(j)$ for $j = 1, \cdots, p$. } $Y = A'(\omega) + \diag[A''(\omega)]^{\sfrac{1}{2}}\ep$, where the coordinates of $\ep$ are conditionally independent and standardized given $\omega$. The covariance of $Y$ is given by the law of total covariance:
\begin{align}
\label{cov}
\C[Y] & = \C[\E(Y|\omega)]  + \E [\C[Y|\omega]] \nonumber \\
& =\C[A'(\omega)] + \E \diag[A''(\omega)].
\end{align}
The off-diagonal entries of the covariance matrix of the noisy images are therefore the same as those of the clean images. However, the diagonal of the covariance matrix (i.e., the variance) of the noisy images is further inflated by the noise variance. 
Unlike white noise, the noise variance here often changes from one coordinate to another (i.e., it is heteroscedastic). In $e$PCA, the homogenization step is a particular weighting method that  improves the signal-to-noise ratio~\cite[Section 4.2.2]{LiuEtAl2017}.  Specifically, the homogenized vector is defined as 
\beq
\label{eq:ePCAh}
Z = \diag[\E A''(\omega)]^{-1/2}Y = \diag[ \E A''(\omega)]^{-1/2} A'(\omega) + \ep.
\eeq
Then the corresponding homogenized covariance matrix becomes
\begin{equation}
\C[Z] =  \diag[\E A''(\omega)]^{-1/2} \C[A'(\omega) ] \diag[\E A''(\omega)]^{-1/2} + I.
\end{equation}
This step is commonly called ``prewhitening'' in signal and image processing, while ``homogenization'' is more commonly used in statistics. The terms ``prewhitened'' and ``homogenized'' are synonyms in the context of the paper. In Section~\ref{sec:hcm}, we discuss how to estimate the homogenized rotationally invariant covariance matrix.

For the special case of Poisson observations $Y\sim \Po_p(X)$, where $X \in \R^p$ is random, we can write $Y = X + \diag(X)^{\sfrac{1}{2}}\ep$. 
The natural parameter is the vector $\omega$ with $\omega(j) = \log X(j)$ and $A'(\omega(j))  = A''(\omega(j))= \exp(\omega(j)) = X(j)$. Therefore, we have $\E Y = \E X $, and 
\begin{equation}
\label{eq:CovPoi}
\C[Y]  =  \C[X] + \diag[ \E X ].
\end{equation}
In words, while the mean of the noisy images agrees with the mean of the clean images, their covariance matrices differ by a diagonal matrix that depends solely on the mean image. If we homogenize the data by $Z = \diag(\E[X])^{-1/2} Y$, then Eq.~\eqref{eq:CovPoi} becomes, 
\begin{align}
    \label{eq:CovZPoi}
        \C[Z] & = \diag(\E[X])^{-1/2} \C[Y] \diag(\E[X])^{-1/2} \nonumber \\ 
        & =  \diag(\E[X])^{-1/2} \C[X] \diag(\E[X])^{-1/2} + I. 
\end{align}
We can estimate the covariance of $X$ from the homogenized covariance $Z$ according to Eq.~\eqref{eq:CovZPoi}, i.e. $\C[X] = \diag(\E[X])^{1/2}(\C[Z] - I)\diag(\E[X])^{1/2}$. In Sec.~\ref{sec:recolor}, we detail the corresponding recoloring step to estimate $\C[X]$. 

In Alg.~\ref{sePCA_alg}, Steps~\ref{a1} and~\ref{a5}, with $D_n = \diag[\bar{Y}]$, correspond to the homogenization procedure in $e$PCA. We provide more details on how to estimate the homogenized rotationally invariant covariance matrix (Steps~\ref{a1}--\ref{a5}) in Sec.~\ref{sec:FB} and Sec.~\ref{sec:hcm}.

\begin{algorithm}[t]
	\SetAlgoLined
	\KwIn{Image data $Y$ that contains $n$ images of size $L \times L$}
	\KwOut{Rotational invariant covariance estimator of noiseless images and denoised images}
	Compute the sample mean $\bar Y = \frac{1}{n}\sum_{i=1}^n Y_i$ {\color{red} ($e$PCA)} \label{a1} \\
    Estimate the support size $R$ and band limit $c$ for the mean image {\color{red} (sPCA)}  \label{a2} \\
	Compute the Fourier-Bessel expansion coefficients of $F(\bar Y)$ and estimate the rotationally invariant sample mean $ \bar{f}$ as in Eq.~\eqref{eq:mean} {\color{red} (s$e$PCA)}  \label{a3}\\
	Compute the variance estimate $D_n  = \diag[\bar{f}]$ {\color{red} (s$e$PCA)}  \label{a4} \\
	Prewhiten the image data $ Z = D_n^{-1/2} Y $ {\color{red} ($e$PCA)}  \label{a5}\\
	Estimate the band limit $c$ for whitened images  {\color{red} (sPCA)}  \label{a6}\\
	Compute the truncated Fourier-Bessel expansion coefficients  of $F(Z)$ and form the coefficients matrices $A^{(k)}$, for $k = 0, \dots, k_{\mathrm{max}}$  {\color{red} (sPCA)}  \label{a7}
	\\	
	\For{$k = 0, 1, \dots, k_{\mathrm{max}}$}{
	Compute the prewhitened sample covariance matrix $S_h^{(k)}$ as in Eq.~\eqref{eq:S_h} and its eigendecomposition $S^{(k)}_{h} = \hat U \Lambda \hat  U^*$ {\color{red} ($e$PCA)}  \label{a9}\\
	Shrink the eigenvalues $S^{(k)}_{h,\eta_{\gamma_k}} = \hat U \eta_{\gamma_k}(\Lambda)\hat U^*$ of top $r_k$ eigenvalues 
	according to Eq.~\eqref{eq:shrinkage} {\color{red} ($e$PCA, sPCA)}  \label{a10}\\
    Compute the recoloring matrix $B^{(k)}$ in eq.~\eqref{eq:Bk} and $D^{(k)}$ in eq.~\eqref{eq:Dk} {\color{red} (s$e$PCA)}  \label{a11}\\
	Recolor the covariance matrix $S^{(k)}_{he} = \left(B^{(k)}\right)^* \cdot S^{(k)}_{h, \gamma_k} \cdot B^{(k)}$ {\color{red} (s$e$PCA)} \label{a12}\\
	Compute the scaling coefficients $\hat \alpha $ in Eq.~\eqref{eq:alpha_def} and keep components with $\hat \alpha > 0$ {\color{red} (s$e$PCA)} \label{a13} \\
	Scale the covariance matrix $S^{(k)}_s=\sum \hat \alpha^{(k)}_i \hat{v}^{(k)}_i \left(\hat{v}^{(k)}_i\right)^*$, where the eigendecomposition of $S^{(k)}_{he}$ is $\sum \hat v^{(k)}_i \left(\hat{v}^{(k)}_i\right)^*$ {\color{red} ($e$PCA)}  \label{a14}\\
    Denoise $\{A^{(k)}\}_{k=0}^{k_{\text{max}}}$ as in Eqs.~\eqref{eq:denoise1} and~\eqref{eq:denoise2} {\color{red} (s$e$PCA)} \label{a15} \\
   }
   The rotational invariant covariance matrix estimator $\hat{\mathcal{S}}((x, y), (x', y')) =  G^{(0)}(x, y) S^{(0)}_s G^{(0)}(x', y')^* + 2 \sum_{k = 1}^{k_\text{max}} G^{(k)}(x, y) S^{(k)}_s G^{(k)}(x', y')^*$.  {\color{red} (sPCA)}\label{a17}\\
   Reconstruct the denoised image using Eq.~\eqref{eq:recon} {\color{red} (sPCA)}\label{a18}
   \caption{Steerable $e$PCA (s$e$PCA) and denoising } \label{sePCA_alg}
\end{algorithm}

\subsection{Steerable basis expansion} 
\label{sec:FB}
Under the observation model in Sec.~\ref{sec:obs_mod}, we develop a method that estimates the rotational invariant $\C[X]$ efficiently and accurately from the image dataset $Y$.  We assume that a digital image $I$ is composed of discrete samples from a continuous function $f$ with band limit $c$. The Fourier transform of $f$, denoted $\mathcal{F}(f)$, can be expanded in any orthogonal basis for the class of squared-integrable functions in a disk of radius $c$. For the purpose of steerable PCA, it is beneficial to choose a basis whose elements are products of radial functions with Fourier angular modes,  
such as the Fourier-Bessel functions, or 2-D prolate functions~\cite{landa2017approximation}. For concreteness, in the following we use the Fourier-Bessel functions given by
\beq
\psi_c^{k, q} (\xi, \theta) = \begin{cases}
N_{k, q} J_k\left(R_{k, q} \frac{\xi}{c} \right) e^{\imath k \theta}, & \xi \leq c, \\
0, & \xi > c,
\end{cases}
\eeq
where $(\xi, \theta)$ are polar coordinates in the Fourier domain (i.e., $\xi_1 = \xi \cos \theta$, $\xi_2 = \xi \sin \theta$, $\xi \geq 0$, and $\theta \in [0, 2\pi)$; $N_{k, q} = (c \sqrt{\pi}|J_{k+1}(R_{k, q})|)^{-1}$ is the normalization factor; $J_k$ is the Bessel function of the first kind of integer order $k$; and $R_{k, q}$ is the $q$th root of the Bessel function $J_k$. We also assume that the functions of interest are concentrated in a disk of radius $R$ in real domain. In order to avoid aliasing, we only use Fourier-Bessel functions that satisfy the following criterion \cite{Klug,zhao2016fast}
\begin{equation}
\label{eq:sampling}
R_{k, q+1} \leq 2\pi c R.
\end{equation}
For each angular frequency $k$, we denote by $p_k$ the number of components satisfying Eq.~\eqref{eq:sampling}. The total number of components is $p = \sum_{k = -k_\text{max}}^{k_\text{max}} p_k $, where $k_{\text{max}}$ is the maximal possible value of $k$ satisfying Eq.~\eqref{eq:sampling}. We also denote $\gamma_k = \frac{p_k}{2n}$ for $k > 0$ and $\gamma_0 = \frac{p_0}{n}$. 

The inverse Fourier transform (IFT) of $\psi_c^{k, q}$ is
\begin{align}
\label{eq:IFT_FB}
\mathcal{F}^{-1}(\psi_c^{k, q})(r, \phi) & = \frac{2 c \sqrt{\pi} (-1)^q R_{k, q} J_k (2 \pi c r)}{\imath^{k}(2\pi c r)^2 - R_{k, q}^2} e^{\imath k \phi} \nonumber \\
&  \equiv g_c^{k, q}(r) e^{\imath k \phi},
\end{align}
where $g_c^{k, q}(r)$ is the radial part of the inverse Fourier transform of the Fourier-Bessel function. 
Therefore, we can approximate $f$ using the truncated expansion
\beq
\label{eq:expansionRD}
f(r, \phi) \approx \sum_{k = -k_\text{max}}^{k_\text{max}} \sum_{q = 1}^{p_k} a_{k, q} g_c^{k, q}(r) e^{\imath k \phi}.
\eeq
The approximation error in Eq.~\eqref{eq:expansionRD} is due to the finite truncation of the Fourier-Bessel expansion. For essentially bandlimited functions, the approximation error is controlled using the asymptotic behavior of the Bessel functions, see~\cite[Section 2]{wang2009rotational} for more details.
We evaluate the Fourier-Bessel expansion coefficients numerically as in \cite{zhao2016fast} using a quadrature rule that consists of equally spaced points in the angular direction $\theta_l = \frac{ 2\pi l }{n_\theta}$, with $l = 0, \dots, n_\theta -1$ and a Gaussian quadrature rule in the radial direction $\xi_j$ for $j = 1, \dots, n_\xi$ with the associated weights $w(\xi_j)$. Using the sampling criterion introduced in~\cite{zhao2016fast}, the values of $n_\xi$ and $n_\theta$ depend on the compact support radius $R$ and the band limit $c$. Our previous work found that using $n_\xi = \left \lceil 4cR \right \rceil$ and $n_\theta = \left \lceil 16cR \right \rceil$ results in highly-accurate numerical evaluation of the integral to evaluate the expansion coefficients.
To evaluate $a_{k, q}$, we need to sample the discrete Fourier transform of the image $I$, denoted $F(I)$, at the quadrature nodes, 
\begin{align}
\label{eq:DFT}
F(I)(\xi_j, \theta_l) & = \frac{1}{2R} \sum_{i_1 = -R}^{R-1}\sum_{i_2 = -R}^{R-1} I(i_1, i_2) \nonumber \\
& \quad \times \exp \left(-\imath 2\pi(\xi_j \cos \theta_l i_1+\xi_j \sin \theta_l i_2)\right),
\end{align}
which can be evaluated efficiently using the the nonuniform discrete Fourier transform~\cite{Greengard}, and we get
\beq
\label{eq:akq}
a_{k, q} \approx \sum_{j = 1}^{n_\xi} N_{k, q}J_{k, q}\left( \frac{\xi_j}{c} \right)\widehat{F(I)}(\xi_j , k)\xi_jw(\xi_j),
\eeq
where $\widehat{F(I)}(\xi_j, k)$ is the 1D FFT of $F(I)$ on each concentric circle of radius $\xi_j$. For real-valued images, it is sufficient to evaluate the coefficients with $k \geq 0$, since $a_{-k, q} = a^*_{k, q}$. In addition, the coefficients have the following properties: under counter-clockwise rotation by an angle $\alpha$, $a_{k, q}$ changes to $a_{k,q}e^{-\imath k \alpha}$; and under reflection, $a_{k, q}$ changes to $a_{-k, q}$. The numerical integration error in Eq.~\eqref{eq:akq} was analyzed in~\cite{zhao2016fast} and drops below $10^{-17}$ for the chosen values of $n_\xi$ and $n_\theta$. 

The steerable basis expansion are applied in two parts of the steerable $e$PCA: (1) the rotationally invariant sample mean estimation in Step~\ref{a3} of Alg.~\ref{sePCA_alg} and (2) the expansion of the whitened images in Step~\ref{a7} of Alg.~\ref{sePCA_alg}.

\subsection{The sample rotationally-invariant homogenized covariance matrix}
\label{sec:hcm}
Suppose $I_1,\ldots,I_n$ are $n$ discretized input images sampled from $f_1,\ldots,f_n$. Here, the observational vectors $Y_i$ for $e$PCA are simply $Y_i = I_i$.
In $e$PCA, the first step is to prewhiten the data using the sample mean as suggested by Eqs.~\eqref{eq:ePCAh} and~\eqref{eq:CovPoi}. However, the sample mean $\bar{Y} = \frac{1}{n} \sum_{i = 1}^n Y_i$ is not necessarily rotationally invariant. With the estimated band limit and support size in Step~\ref{a2}, we compute the truncated Fourier-Bessel expansion coefficients of $F(\bar{Y})$, denoted by  $\bar a_{k, q}$. The rotationally invariant sample mean can be evaluated from $\bar a_{k, q}$,
\beq
\label{eq:mean}
\bar{f}(r, \phi) = \frac{1}{2\pi}\int_0^{2\pi} \frac{1}{n} \sum_{i = 1}^n  f_i(r, \phi - \alpha) d \alpha \approx \sum_{q=1}^{p_0} \bar{a}_{0, q} g_{c}^{0, q}(r).
\eeq
The approximation error in Eq.~\eqref{eq:mean} follows directly from that of Eq.~\eqref{eq:expansionRD}.
The rotationally invariant sample mean is circularly symmetric.
We denote by $\bar{A}$ a vector that contains all the coefficients $\bar{a}_{0, q}$ ordered by the radial index $q$. Although the input images are non-negative, the finite truncation may result in small negative values in the mean estimation, so we threshold any negative entries to zero. 
As mentioned in Sec.~\ref{sec:obs_mod}, the covariance matrix of the Poisson observations differs by a diagonal matrix, where the diagonal entries are equal to the mean image. Therefore, in Step~\ref{a4} of Alg.~\ref{sePCA_alg}, we have $D_n = \diag[\bar{f}]$ and it is used in Step~\ref{a5} to compute the homogenized vectors, similar to Eq.~\eqref{eq:ePCAh}.

We prewhiten the images by the estimated mean image to create new images $Z_1,\ldots,Z_n$ as $Z_i (x, y) = \bar{f}(x, y)^{-1/2} Y_i(x, y)$, when $\bar{f}(x, y) > 0$, and $Z_i(x, y)$ is 0 otherwise.  The whitening step might change the band limit. Therefore, we estimate the band limit of the whitened images in Step~\ref{a6}. 
Combining Eqs.~\eqref{eq:sampling} and~\eqref{eq:akq}, we compute the truncated Fourier-Bessel expansion coefficients $a_{k,q}^i$ of $F(Z_i)$. Let us denote by $A^{(k)}$ the matrix of expansion coefficients with angular frequency $k$, obtained by putting $a^i_{k, q}$ into a matrix, where the columns are indexed by the image number $i$ and the rows are ordered by the radial index $q$. The coefficient matrix $A^{(k)}$ is of size $p_k \times n$.

We use $\bar{Z}$ to represent the rotational invariant sample mean of the whitened images.   Under the action of the group $O(2)$, i.e. counter-clock wise rotation by an angle $\alpha \in [0, 2\pi)$ and reflection $\beta \in \{+, -\}$, where `$+$' indicates no reflection and `$-$' indicates with reflection,  the image $Z_i$ is transformed to $Z_i^{\alpha, \beta}$. Since the truncated Fourier-Bessel transform is almost unitary~\cite{zhao2016fast}, the rotationally invariant covariance kernel $\mathcal{S}\left((x, y),(x', y')\right)$ built from the whitened image data with all possible in-plane rotations and reflections, defined as, 
\begin{align}
   &\mathcal{S}\left((x, y),(x', y')\right)  = \frac{1}{ 4\pi n } \sum_{i = 1}^n \sum_{\beta \in \{+, -\}}\int_0^{2\pi} \left ( Z^{\alpha, \beta}_i(x, y) \right. \nonumber \\
 & \quad \quad \quad  \left. - \bar{Z}(x, y) \right )  \left ( Z^{\alpha, \beta}_i(x', y') - \bar{Z}(x', y') \right )  d\alpha,
       \label{eq:cov_kernel}
\end{align}
can be computed in terms of the IFT of the Fourier-Bessel basis and the associated expansion coefficients. Subtracting the sample mean is equivalent to subtracting $\frac{1}{n}\sum_{j=1}^n a_{0,q}^j $ from the coefficients $a^i_{0, q}$, while keeping other coefficients unchanged. Therefore, we first update the zero angular frequency coefficients by $a^i_{0, q} \gets a^i_{0, q} - \frac{1}{n}\sum_{j=1}^n a_{0,q}^j$. In terms of the expansion coefficients, the rotational invariant homogenized sample covariance matrix is $S_h = \bigoplus_{k = -k_{\text{max}}}^{k_\text{max}} S_h^{(k)}$, with
\beq
\label{eq:S_h}
S_h^{(k)} = \frac{1}{n} \text{Re}\left\{ A^{(k)} \left(A^{(k)}\right)^*\right \}.
\eeq
We further denote the eigenvalues and eigenvectors of $S_h^{(k)}$ by $\lambda_i^{(k)}$ and $\hat{u}^{(k)}_i$, that is,
\beq
S_h^{(k)} = \sum_{i=1}^{p_k} \lambda^{(k)}_i \hat{u}^{(k)}_i (\hat{u}^{(k)}_i)^*.
\eeq

The procedure of homogenization for steerable $e$PCA is detailed in Steps~\ref{a1}--\ref{a5} in Alg.~\ref{sePCA_alg}. Step~\ref{a9} of Alg.~\ref{sePCA_alg} computes the rotationally invariant homogenized sample covariance matrix. 

\subsection{Eigenvalue shrinkage}
\label{sec:shrinkage}
For data corrupted by additive white noise (with noise variance 1 in each coordinate), previous works~\cite{baik2005phase, baik2006eigenvalues,paul2007asymptotics, benaych2011eigenvalues} show that if the population eigenvalue $\ell$ is above the Baik, Ben Arous, P\'ech\'e (BBP) phase transition, then the sample eigenvalue pops outside of the Mar{\v c}enko-Pastur distribution of the ``noise'' eigenvalues. The  sample eigenvalue will converge to the value given by \emph{the spike forward map} as $p_k, n \to \infty$, and $p_k/n=\gamma_k$:
\begin{equation}
\label{eq:spikefmap}
\lambda(\ell;\gamma_k)=
\left\{
	\begin{array}{ll}
		(1+\ell) \left(1+ \frac{\gamma_k}{\ell}\right) & \mbox{\, if \, } \ell> \sqrt{\gamma_k}, \\
		\left(1+ \sqrt{\gamma_k} \right)^2 & \mbox{\, otherwise.}
	\end{array}
\right.
\end{equation}
The underlying clean population covariance eigenvalues can be estimated by solving the quadratic equation in Eq.~\eqref{eq:spikefmap},
\begin{align}
\label{eq:shrinkage}
& \hat{\ell} = \eta_{\gamma_k}(\lambda)  \nonumber \\
& = \begin{cases}
\frac{\left(\lambda - 1 - \gamma_k + \sqrt{(\lambda - 1 - \gamma_k)^2 - 4\gamma_k}\right)}{2} , & \lambda > (1 + \sqrt{\gamma_k})^2 , \\
0 &  \lambda \leq (1 + \sqrt{\gamma_k})^2 .
\end{cases}
\end{align}
Shrinking the eigenvalues improves the estimation of the sample covariance matrix~\cite{Donoho2013}. Since the homogenized sample covariance matrix $S_h$ is decoupled into small sub-blocks $S_{h}^{(k)}$, the shrinkers are defined for each frequency $k$ separately. 
The shrinkers $\eta_{\gamma_k}(\lambda)$ set all noise eigenvalues to zero for $\lambda$ within the support of the Mar{\v c}enko-Pastur distribution and reduce other eigenvalues according to Eq.~\eqref{eq:shrinkage}. 
Then the denoised covariance matrices are
\beq
\label{eq:S_heta}
S_{h, \eta}^{(k)} = \sum_{i = 1}^{r_k} \eta_{\gamma_k}\left(\lambda^{(k)}_i \right) \hat{u}^{(k)}_i (\hat{u}^{(k)}_i)^*,
\eeq
where $r_k$ is the number of components with $\eta_{\gamma_k} (\lambda) > 0$.
The empirical eigenvector $\hat{u}^{(k)}$ of $S^{(k)}_h$ is an inconsistent estimator of the true eigenvector. We can heuristically quantify the inconsistency based on results from the Gaussian standard spiked model, even though the noise is non-Gaussian. Under this model, the empirical and true eigenvectors have an asymptotically deterministic angle: $(\left(u^{(k)}\right)^* \hat u^{(k)})^2 \to c^2(\ell;\gamma_k)$ almost surely, where $c(\ell;\gamma_k)$ is the \emph{cosine forward map} given by \cite{paul2007asymptotics, benaych2011eigenvalues}:
\beq
c^2(\ell;\gamma_k)=
\left\{
	\begin{array}{ll}
		\frac{1-\gamma_k/\ell^2}{1+\gamma_k/\ell} & \mbox{\, if \, } \ell>\sqrt{\gamma_k}, \\
		0 &  \mbox{\, otherwise.}
	\end{array}
\right.
\eeq
Therefore, asymptotically for the population eigenvectors beyond the BBP phase transition, the sample eigenvectors have positive correlation with the population eigenvectors, but this correlation is less than 1~\cite{johnstone2009consistency,jung2009pca,johnstone2018pca}. We denote by $\hat{c}$ an estimate of $c$ using the estimated clean covariance eigenvalues $\hat \ell$ in Eq.~\eqref{eq:shrinkage} and $\hat{s}^2 = 1 - \hat{c}^2$. 

In short, Step~\ref{a10} of Alg.~\ref{sePCA_alg} improves the estimation of the rotationally invariant homogenized covariance matrix through eigenvalue shrinkage. 

\begin{figure}
\captionsetup[subfloat]{farskip=2pt,captionskip=1pt}
	\begin{center}
		\subfloat[clean]{
			\includegraphics[width = 0.45\columnwidth]{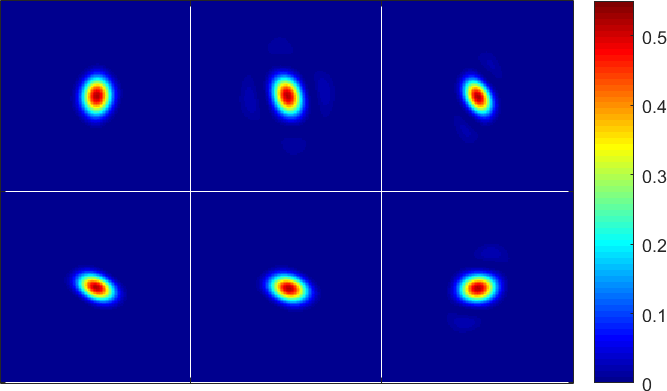}
			\label{fig:clean_images}
		}
		\subfloat[noisy]{
			\includegraphics[width = 0.45\columnwidth]{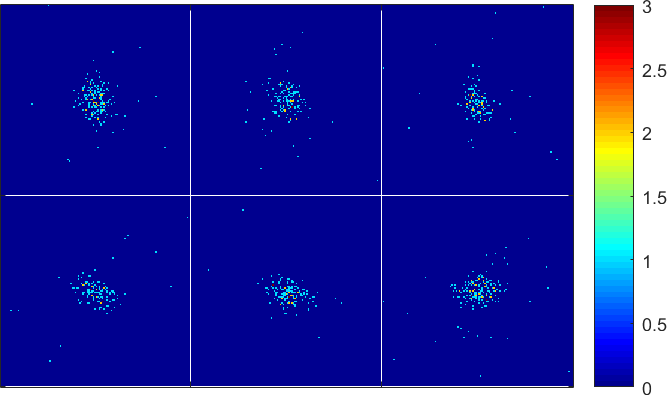}
			\label{fig:noisy_images}
		} 
	\end{center}
	\caption{Sample clean and noisy images of the XFEL dataset. Image size is $128 \times 128$ with mean per pixel photon count$=0.01$.}
	\label{fig:data}
\end{figure} 

\begin{figure}
\vspace{-0.5cm}
 \captionsetup[subfigure]{aboveskip=-1pt,belowskip=-1pt}
    \centering
    \subfloat[Estimate $R$]{
    \includegraphics[width = 0.45 \columnwidth]{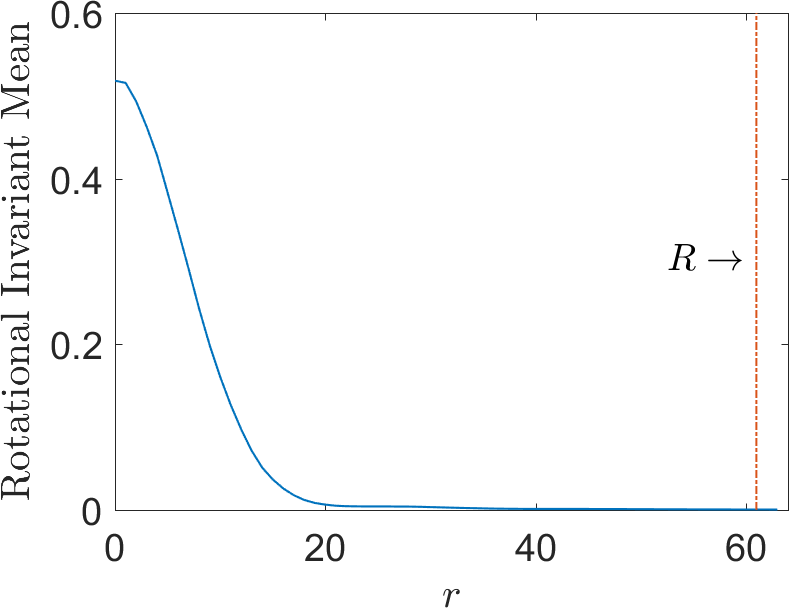}
    \label{fig:R}
    }
    \subfloat[Estimate $c$]{
    \includegraphics[width = 0.45 \columnwidth]{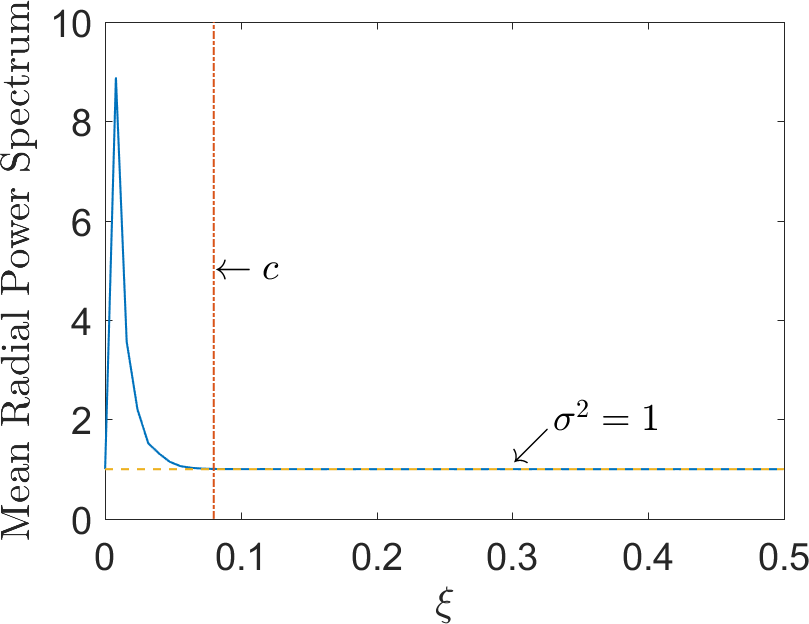}
    \label{fig:c}
    }
    \caption{Estimating $R$ and $c$ from $n = 7\times 10^4$ simulated noisy XFEL diffraction intensity maps of lysozyme. Each image is of size $128 \times 128$ pixels. \protect \subref{fig:R} The radial profile of the rotational invariant sample mean image. The radius of compact support is chosen at $R = 61$. \protect \subref{fig:c} Mean radial power spectrum of the whitened noisy images. The curve levels off at $\sigma^2 = 1$. }
    \label{fig:cR}
\end{figure}

\subsection{Recoloring}
\label{sec:recolor}
Homogenization changes the direction of the clean eigenvectors. Therefore, after eigenvalue shrinkage, we recolor (heterogenize) the covariance matrix by conjugating the recoloring matrix $B$ with $S_{h, \eta}$: $S_{he} = B^* \cdot S_{h, \eta} \cdot B$. The recoloring matrix is derived as,
\begin{align}
B_{k_1, q_1; k_2, q_2} & = \int_0^R \int_0^{2\pi} \sqrt{\bar{f}( r )}\, \mathcal{F}^{-1}\left(\psi_c^{k_1, q_1}\right)(r, \theta)\,  \nonumber \\
& \quad \quad \quad \times \left(\overline{\mathcal{F}^{-1}\left(\psi_c^{k_2, q_2}\right)(r, \theta)}\right)\, r dr  d\theta  \nonumber \\
& = \int_0^R \sqrt{\bar{f}(r)}\, g_c^{k_1, q_1}\left( r \right)\, \overline{g_c^{k_2, q_2}\left(r \right)} \, r d r \nonumber \\
& \quad \quad  \times \int_0^{2\pi} e^{\imath (-k_1 + k_2) \theta} d\theta \nonumber \\
& = \delta_{k_1, k_2} \int_0^R \sqrt{\bar{f}(r)} g_c^{k_1, q_1}\left( r \right) \overline{g_c^{k_2, q_2}\left(r\right)} r d r,
\end{align}
which has a block diagonal structure and is decoupled for each angular frequency, $B = \bigoplus_{k = -k_{\text{max}}}^{k_{\text{max}}} B^{(k)}$, with 
\beq
\label{eq:Bk}
B^{(k)}_{q_1, q_2} = \int_0^R \sqrt{\bar{f}(r)}\, g_c^{k, q_1}\left( r \right)\, \overline{g_c^{k, q_2}\left(r\right)}\, r d r .
\eeq
The radial integral in Eq.~\eqref{eq:Bk} is numerically evaluated using the Gauss-Legendre quadrature rule~\cite[Chap. 4]{press1996numerical}, which determines the locations of $n_r = \lceil 4cR \rceil$ points $\{r_j\}_{j = 1}^{n_r}$ on the interval $[0, R]$ and the associated weights $w(r_j)$. The integral in Eq.~\eqref{eq:Bk} is thus approximated by
\begin{equation}
    \label{eq:B_nint}
B^{(k)}_{q_1, q_2} \approx \sum_{j = 1}^{n_r} \sqrt{\bar{f}(r_j)}\, g_c^{k, q_1}\left( r_j \right)\, \overline{g_c^{k, q_2}\left(r_j \right)}r_j w(r_j).  
\end{equation}

\begin{figure*}
\vspace{-0.5cm}
\captionsetup[subfloat]{farskip=2pt,captionskip=1pt}
	\subfloat[]{	
		\includegraphics[width = 0.19\textwidth]{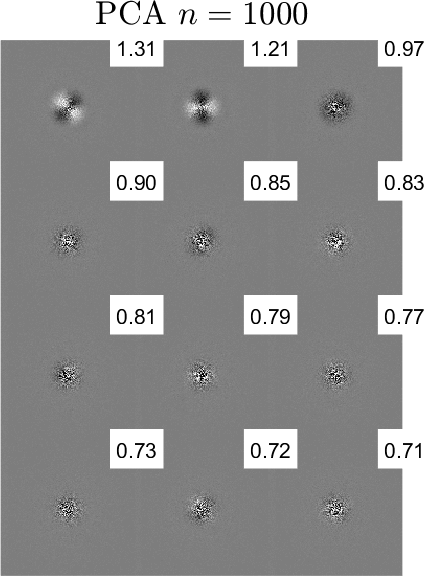}%
		\label{fig:pc_pca1000}
	}%
	\subfloat[]{	
		\includegraphics[width = 0.19\textwidth]{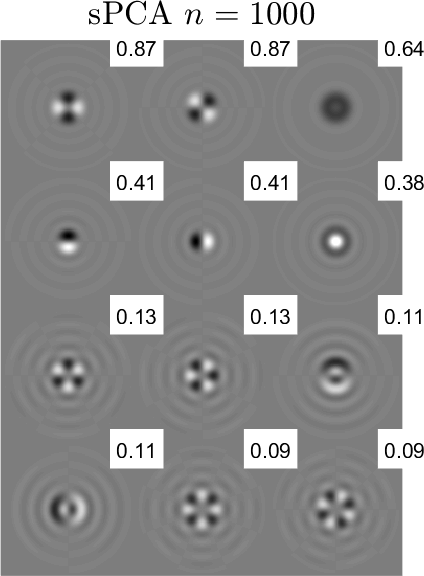}%
		\label{fig:pc_spca1000}
	}%
	\subfloat[]{	
		\includegraphics[width = 0.19\textwidth]{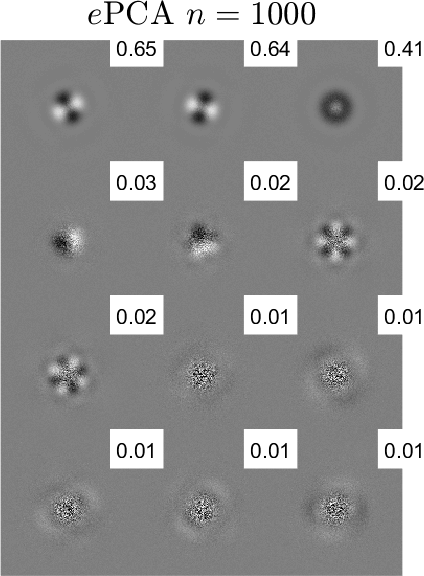}%
		\label{fig:pc_epca1000}
	}%
	\subfloat[]{	
		\includegraphics[width = 0.19\textwidth]{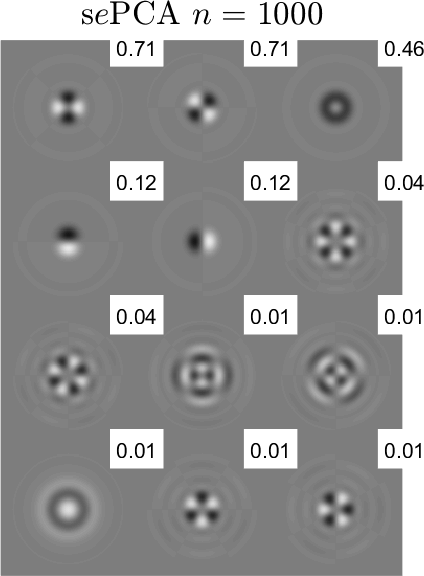}%
		\label{fig:pc_sepca1000}
	} 	
	\subfloat[]{	
	\includegraphics[width = 0.19\textwidth]{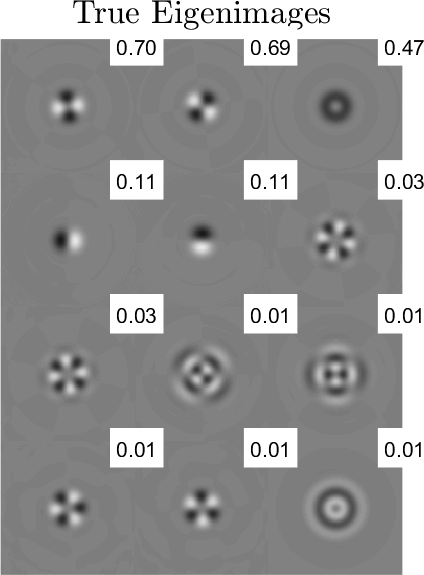}%
	\label{fig:true_pc}
}%
\\ 
	\subfloat[]{	
		\includegraphics[width = 0.19\textwidth, trim=0 0 0 0, clip]{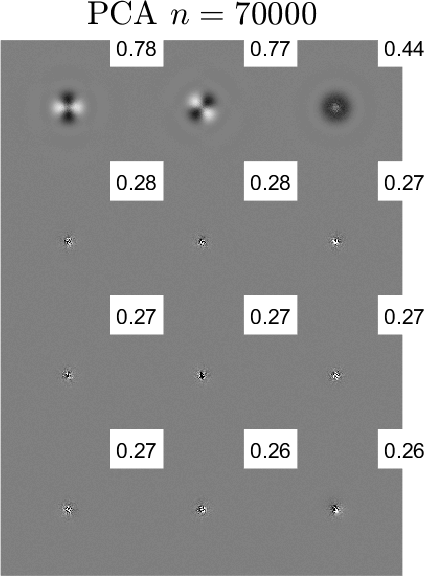}
		\label{fig:pc_pca70000}
	}%
	\subfloat[]{	
		\includegraphics[width = 0.19\textwidth]{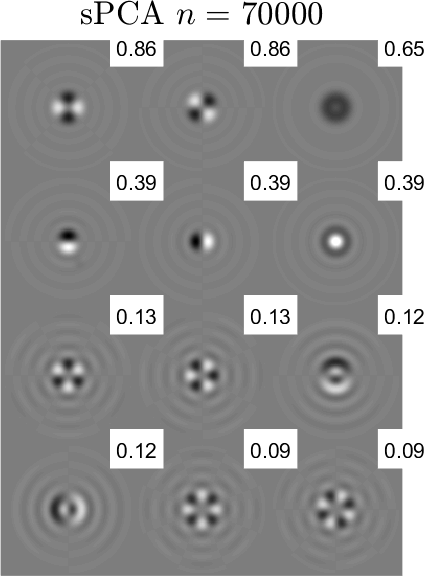}
		\label{fig:pc_spca70000}
	}%
	\subfloat[]{
		\includegraphics[width = 0.19\textwidth]{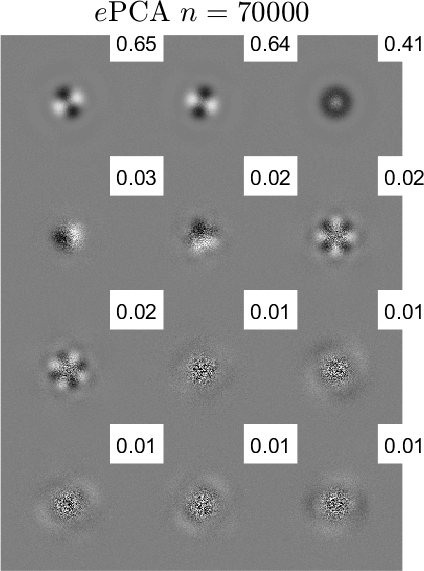}
		\label{fig:pc_epca70000}
	}%
	\subfloat[]{	
		\includegraphics[width = 0.19\textwidth]{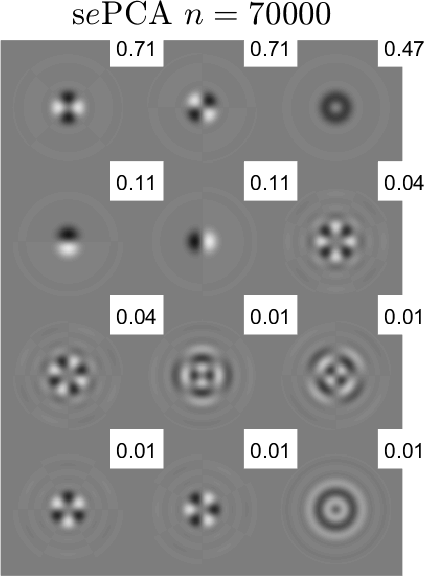}
		\label{fig:pc_sepca70000}
	}	
	\caption{Eigenimages estimated from noisy XFEL data using PCA, steerable PCA (sPCA), $e$PCA, and steerable $e$PCA (s$e$PCA), ordered by eigenvalues. Input images are corrupted by Poisson noise with mean photon count 0.01 (shown in Figure \ref{fig:noisy_images}). } 
	\label{fig:eigVec}
	\vspace{-0.5cm}
\end{figure*}

The recoloring step is also decoupled for each angular frequency sub-block. The heterogenized covariance estimators are
\beq
\label{eq:S_he}
S^{(k)}_{he} = \left(B^{(k)}\right)^* \cdot S^{(k)}_{h, \eta_{\gamma_k}} \cdot B^{(k)}.
\eeq
Similar to Eq.~\eqref{eq:Bk}, we define $D^{(k)}$ which will be used to scale the heterogenized covariance matrix estimator (see Eq.~\eqref{eq:alpha_def}) and denoise the expansion coefficients (see Eqs.~\eqref{eq:denoise1} and~\eqref{eq:denoise2}),
\begin{align}
D^{(k)}_{q_1, q_2} & = \int_0^R \bar{f}(r)\, g_c^{k, q_1}\left( r \right)\, \overline{g_c^{k, q_2}\left(r\right)}\, r d r \nonumber \\
& \approx \sum_{j = 1}^{n_r} \bar{f}(r_j)\, g_c^{k, q_1}\left( r_j \right)\, \overline{g_c^{k, q_2}\left(r_j\right)} r_j w(r_j).
\label{eq:Dk}
\end{align}
In Alg.~\ref{sePCA_alg}, Steps~\ref{a11} and~\ref{a12}  summarize the recoloring procedure.   

\subsection{Scaling} 
\label{sec:scaling}
The eigendecomposition of $S^{(k)}_{he}$ gives $S^{(k)}_{he} = \sum_{i = 1}^{r_k} \hat{v}_i^{(k)} \left( \hat{v}_i^{(k)} \right)^*$. The empirical eigenvalues are $\hat t = \| \hat{v}^{(k)} \|^2$ which is a biased estimate of the true eigenvalues of the clean covariance matrix $\Sigma_X$.
In~\cite[Sec. 4.2.3]{LiuEtAl2017}, a scaling rule was proposed to correct the bias. We extend it in Steps~\ref{a13} and~\ref{a14} in Alg.~\ref{sePCA_alg} to the steerable case and scale each eigenvalue of $S^{(k)}_{he}$ by a parameter $\hat{\alpha}^{(k)}$,
\beq
\label{eq:alpha_def}
\hat \alpha^{(k)}  = \begin{cases}
\frac{1 - \hat{s}^2 \tau^{(k)}}{ \hat{c}^2 }, & \text{for} \quad 1 - \hat{s}^2 \tau^{(k)} > 0\,\;  \text{ and }\,\; \hat{c}^2 > 0 \\
0 & \text{otherwise}
\end{cases}
\eeq
where the parameter $\tau^{(k)} = \frac{\tr{D^{(k)}}}{p_k} \cdot \frac{\hat \ell}{\|\hat v^{(k)} \|^2 }$. The scaled covariance matrices are
\beq
\label{eq:scaled_cov}
S^{(k)}_s = \sum_{i=1}^{r_k} \hat{\alpha}^{(k)}_i \hat v^{(k)}_i \left(\hat{v}^{(k)}_i\right)^*.
\eeq
The rotational invariant covariance kernel $\mathcal{S}((x, y), (x', y') )$ is well approximated by $ \sum_{k = -k_{\text{max}}}^{k_{\text{max}}} G^{(k)}(x, y) S^{(k)}_s \left(G^{(k)}(x', y')\right)^*$, where $G^{(k)}$ contains IFT of all $\psi_c^{k, q}$ with angular frequency $k$ (see Step~\ref{a17}) in Alg.~\ref{sePCA_alg}). The computational complexity of steerable $e$PCA is $O(n L^3 + L^4)$, same as steerable PCA, and it is lower than the complexity of $e$PCA which is $O(\min(nL^4 + L^6, n^2L^2 + n^3))$.

In summary, Step~\ref{a14} scales the heterogenized covariance matrix $S_{he}$. The covariance matrix in the original pixel domain is efficiently computed from $S_s^{(k)}$ in Step~\ref{a17}.  

\begin{figure*}
\captionsetup[subfloat]{farskip=2pt,captionskip=1pt}
	\centering
	\subfloat[]{
		\includegraphics[width = 0.4 \textwidth]{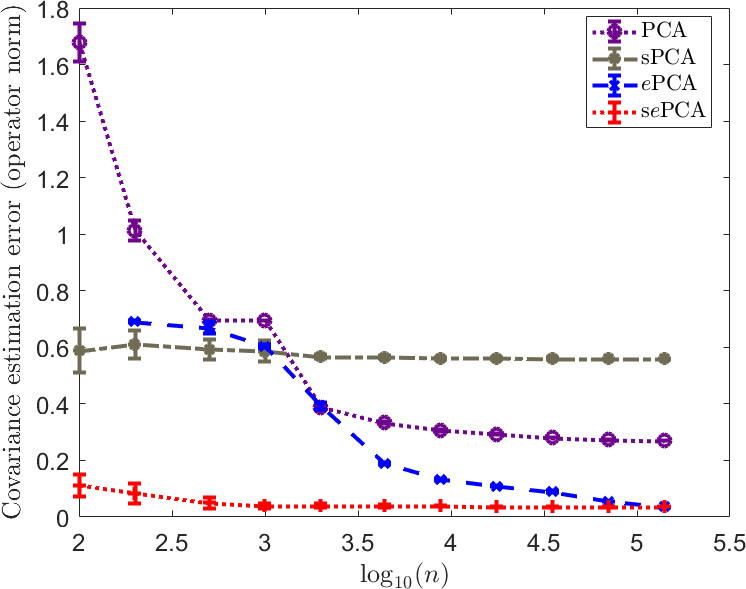}
		\label{fig:OpNorm}
	} 
	\subfloat[]{
		\includegraphics[width = 0.4 \textwidth]{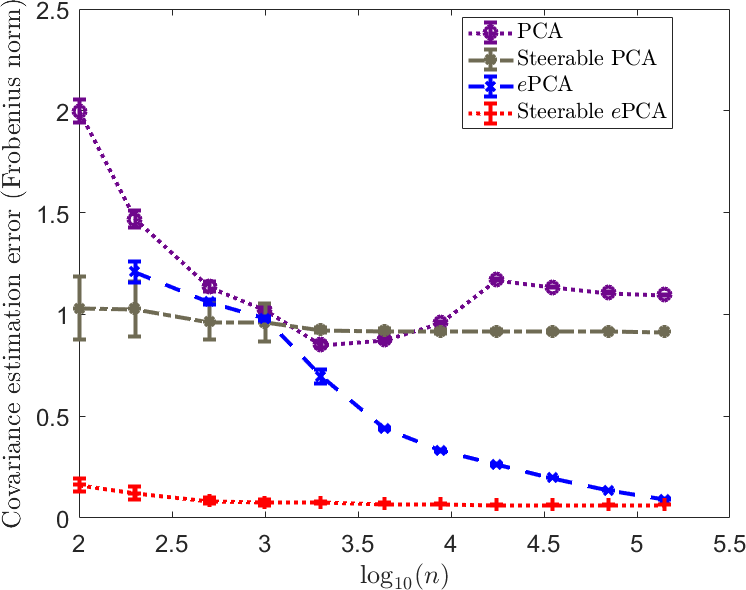} 
		\label{fig:FroNorm}
	}
	\caption{Error of covariance matrix estimation, measured as the \protect \subref{fig:OpNorm} operator norm and \protect \subref{fig:FroNorm} Frobenius norm of the difference between each covariance estimate and the true covariance matrix. The sample size $n$ ranges from 100 to 140,000.}
	\label{fig:xfel_err_cov_est}
\end{figure*}

\subsection{Denoising}
As an application of steerable $e$PCA, we develop a method to denoise the photon-limited images. Since the Fourier-Bessel expansion coefficients are computed from the prewhitened images, we first recolor the coefficients by multiplying $B^{(k)}$ with $A^{(k)}$ and then we apply Wiener-type filtering to denoise the coefficients. For the steerable basis expansion coefficients $A^{(k)}$ with angular frequency $k \neq 0$,
\beq
\label{eq:denoise1}
\hat{A}^{(k)} = S^{(k)}_s \left( D^{(k)} + S^{(k)}_s \right)^{-1} B^{(k)} A^{(k)}.
\eeq
For $k = 0$, we need to take into account the rotational invariant mean expansion coefficients,
\begin{align}
\hat{A}^{(0)} &= S^{(0)}_s \left( D^{(0)} + S^{(0)}_s \right )^{-1} B^{(0)} A^{(0)} \nonumber \\
& \quad + D^{(0)} \left( D^{(0)} + S^{(0)}_s \right)^{-1} \bar{A} \mathbf 1_{n}^\top.
\label{eq:denoise2}
\end{align}
The denoised image sampled on the Cartesian grid $(x, y)$ in real domain are computed from the filtered expansion coefficients $\hat{a}_{k,q}^i$,
\begin{align}
\label{eq:recon}
\hat{X}_i (x, y)  & = \sum_{q = 1}^{p_0} \hat{a}_{0, q}^i \, g_c^{0, q}(r_{x, y}) \nonumber \\ 
& \quad + 2 \text{Re}\left [ \sum_{k = 1}^{k_{\text{max}}} \sum_{q = 1}^{p_k} \hat{a}_{k, q}^i \, g_c^{k, q}(r_{x, y}) e^{-\imath k \theta_{x, y}} \right ],
\end{align}
where $r_{x, y} = \sqrt{x^2 + y^2}$ and $\theta_{x, y} = \tan^{-1}\left( \frac{y}{x}\right)$.

\begin{figure}
    \centering
    \includegraphics[width=0.8\columnwidth]{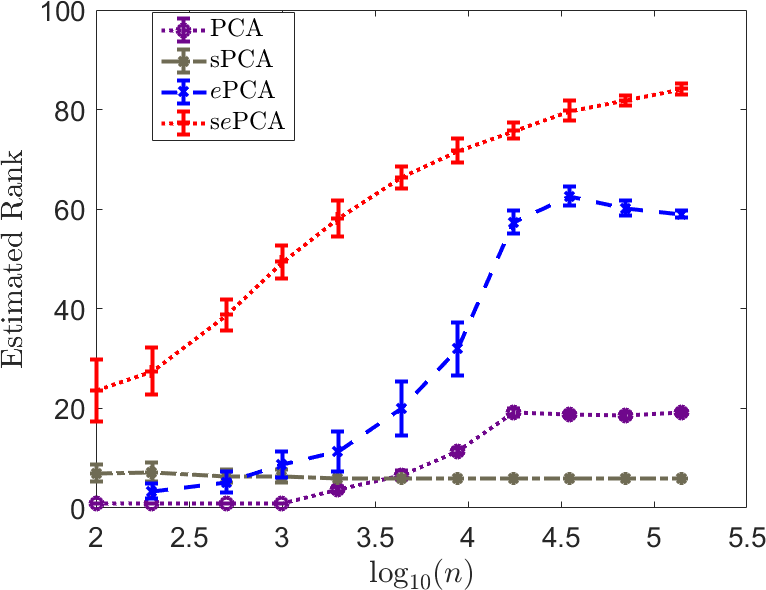}
        \caption{The estimated number of signal principal components using PCA, sPCA, $e$PCA and s$e$PCA.  Images are corrupted by Poisson noise with mean per pixel photon count 0.01.}
        \label{fig:estimated_ranks}
\end{figure}

In essense, we first denoise the recolored steerable expansion coefficients in Step~\ref{a15} of Alg.~\ref{sePCA_alg} according to Eqs.~\eqref{eq:denoise1} and~\eqref{eq:denoise2} and then reconstruct the image using Eq.~\eqref{eq:recon} in Step~\ref{a18} of Alg.~\ref{sePCA_alg}. 

\begin{figure}
    \centering
    \includegraphics[width=0.8\columnwidth]{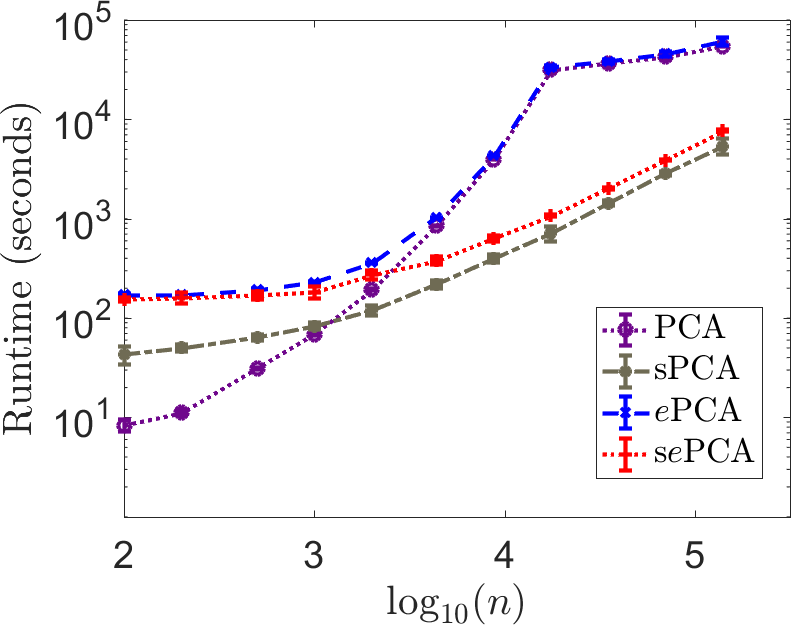}
    \caption{The runtimes for computing the principal components using PCA, sPCA, $e$PCA and s$e$PCA. Images are corrupted by Poisson noise with mean per pixel photon count $=0.01$. }
    \label{fig:runtime}
\end{figure}

\begin{figure}
	\begin{center}
	\subfloat[Clean]{
		\includegraphics[width = 0.48\columnwidth]{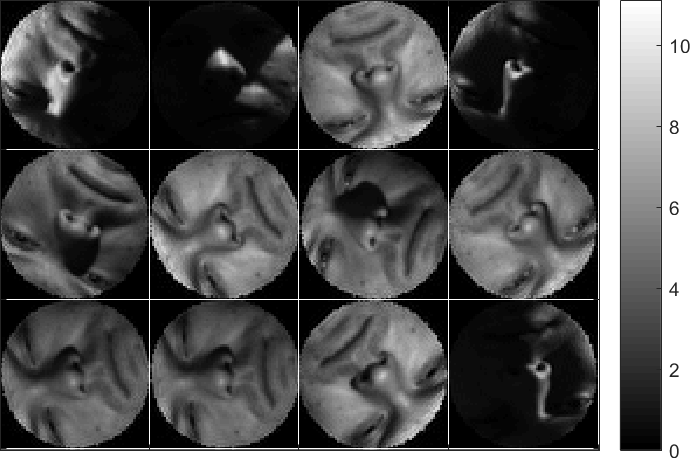}
		\label{fig:face_clean}
		}
	\subfloat[Noisy]{
		\includegraphics[width = 0.48\columnwidth]{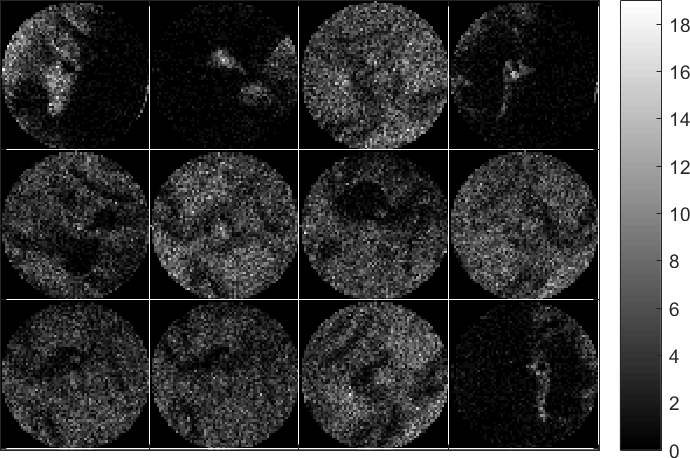}
		\label{fig:face_noisy}
	}
	\end{center}
	\caption{Sample clean and noisy images of randomly rotated Yale B face data. Image size is $64 \times 64$ with mean intensity 2.3 photons per pixel. }
	\label{fig:face_data}
\end{figure}

\begin{figure*}[h!]
\vspace{-0.5cm}
\captionsetup[subfloat]{farskip=2pt,captionskip=1pt}
	\subfloat[]{	
		\includegraphics[width = 0.19\textwidth]{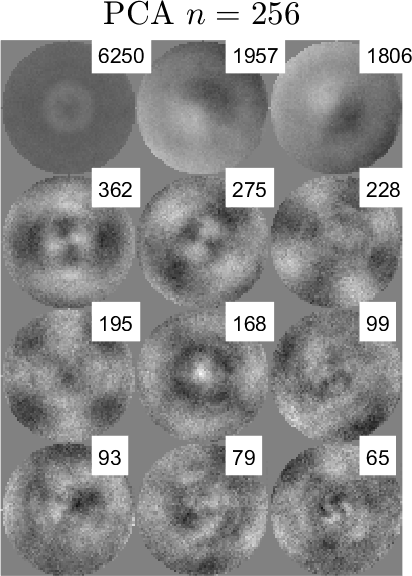}%
		\label{fig:pc_pca256}
	}%
	\subfloat[]{	
		\includegraphics[width = 0.19\textwidth]{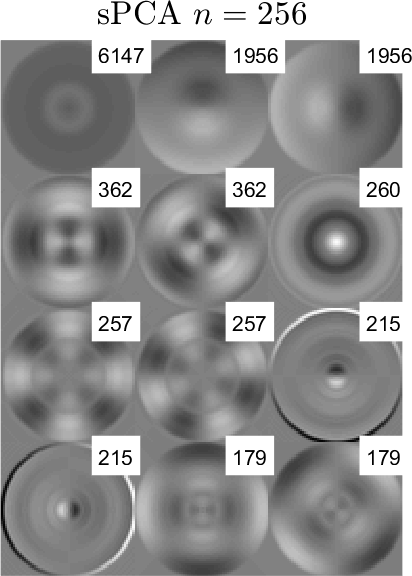}%
		\label{fig:pc_spca256}
	}%
	\subfloat[]{	
		\includegraphics[width = 0.19\textwidth]{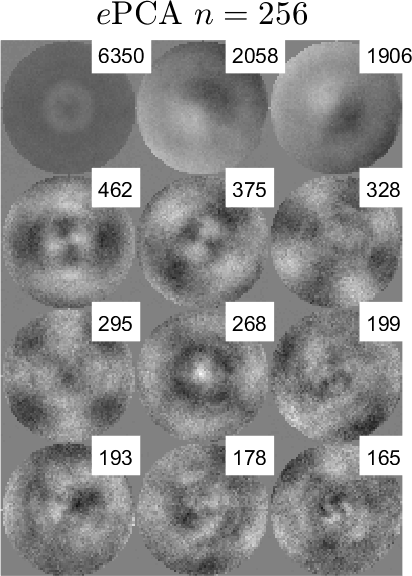}%
		\label{fig:pc_epca256}
	}%
	\subfloat[]{	
		\includegraphics[width = 0.19\textwidth]{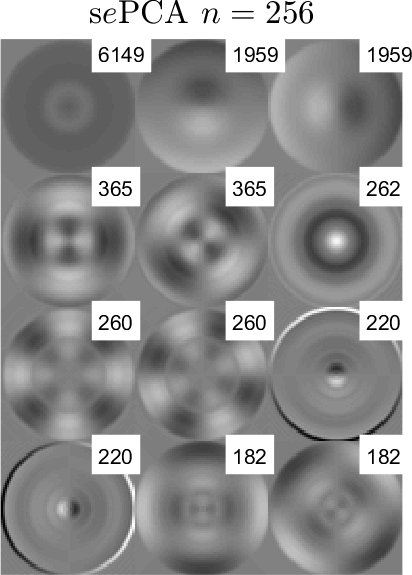}%
		\label{fig:pc_sepca256}
	} 	
	\subfloat[]{	
	\includegraphics[width = 0.19\textwidth]{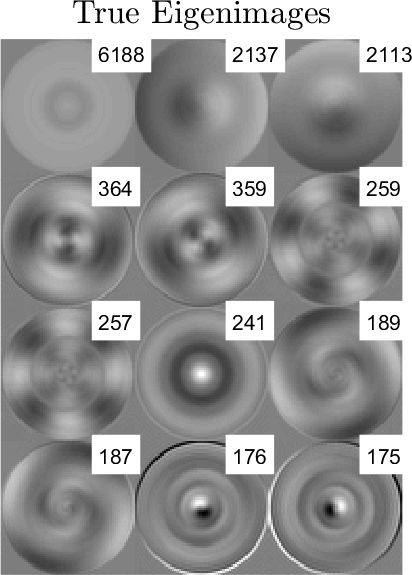}%
	\label{fig:true_pc_face}
}%
\\ 
	\subfloat[]{	
		\includegraphics[width = 0.19\textwidth, trim=0 0 0 0, clip]{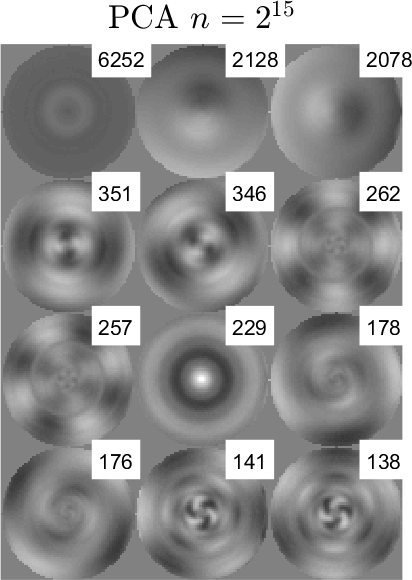}
		\label{fig:pc_pca32768}
	}%
	\subfloat[]{	
		\includegraphics[width = 0.19\textwidth]{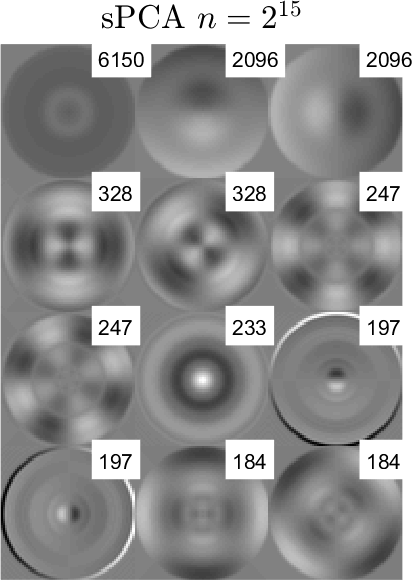}
		\label{fig:pc_spca32768}
	}%
	\subfloat[]{
		\includegraphics[width = 0.19\textwidth]{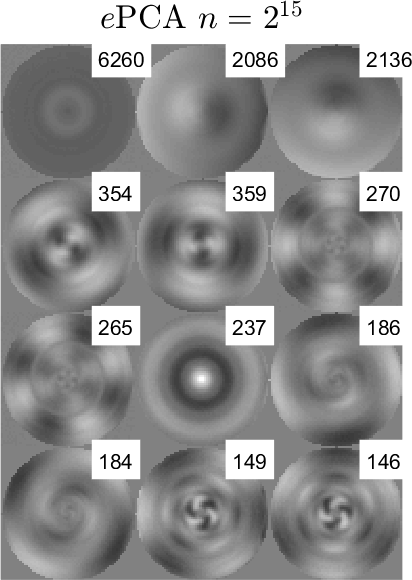}
		\label{fig:pc_epca32768}
	}%
	\subfloat[]{	
		\includegraphics[width = 0.19\textwidth]{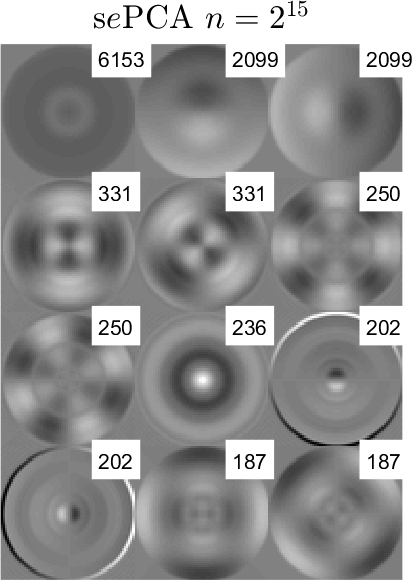}
		\label{fig:pc_sepca32768}
	}	
	\caption{Eigenimages estimated from noisy rotated Yale B face data using PCA, sPCA, $e$PCA, and s$e$PCA, ordered by eigenvalues. Input images are corrupted by Poisson noise with mean intensity 2.3 photons per pixel (shown in Fig. \ref{fig:face_noisy}). } 
	\label{fig:eigVec_face}
	\vspace{-0.5cm}
\end{figure*}

\section{Numerical Results}
\label{sec:results}
We apply PCA, $e$PCA, steerable PCA, and steerable $e$PCA to a simulated XFEL dataset and compare the results for covariance estimation and denoising. The algorithms are implemented in MATLAB on a machine with 60 cores, running at 2.3 GHz, with total RAM of 1.5TB. Only 8 cores were used in our experiments. 

We simulate $n = 140,000$ noiseless XFEL diffraction intensity maps of a lysozyme (Protein Data Bank 1AKI) with Condor~\cite{maia2016condor}.
The average pixel intensity is rescaled to be 0.01 for image size $128 \times 128$  pixels such that shot noise dominates~\cite{SchwanderEtAl2012}. To sample an arbitrary number $n$ of noisy diffraction patterns, we sample an intensity map at random, 
and then sample the photon count of each detector pixel from a Poisson distribution whose mean is the pixel intensity. Figs.~\ref{fig:clean_images} and~\ref{fig:noisy_images} illustrate the clean intensity maps and the resulting noisy diffraction patterns. 

We estimate the radius of the concentration of the diffraction intensities in real domain and the band limit in Fourier domain from the noisy images in the following way. The data variance is proportional to the sample mean. Therefore, we estimate the rotationally invariant sample mean and show the radial profile in Fig.~\ref{fig:R}, which is also the variance map of the dataset averaged in the angular direction. At large $r$, the radial part of the rotational invaraint sample mean levels off at $0$.  We compute the cumulative variance by integrating the radial sample mean over $r$ with a Jacobian weight $rdr$. The fraction of the cumulative mean exceeds 99.9\% at $r = 61$, and therefore $R$ was chosen to be $61$ (see Fig.~\ref{fig:R}). 
We compute the whitened projection images using the rotational invariant sample mean. For the whitened images, we compute the angular average of the mean 2D power spectrum. The curve in Fig.~\ref{fig:c} levels off at
the noise variance $\sigma^2 = 1$ when $\xi$ is large. We used the same method as before to compute the cumulative radial power spectrum. The fraction reaches 99.9\% at $\xi =  0.08$, therefore the band limit is chosen to be $c = 0.08$. The band limit $c$ and support radius $R$ are used in both steerable PCA and steerable $e$PCA. 

\subsection{Covariance estimation and principal components}
Fig.~\ref{fig:eigVec} shows the top 12 eigenimages for clean XFEL diffraction patterns (Fig.~\ref{fig:true_pc}), and noisy diffraction patterns with mean photon count per pixel 0.01 (Figs.~\ref{fig:pc_epca1000}--\ref{fig:pc_sepca70000}) using PCA, steerable PCA, $e$PCA, and steerable $e$PCA. The true eigenimages in Fig.~\ref{fig:true_pc} are computed from 70000 clean diffraction patterns whose orientations are uniformly distributed over $SO(3)$. Figs.~\ref{fig:pc_pca1000}--\ref{fig:pc_sepca1000} are computed from $1000$ noisy images. Since the number of samples is much smaller than the size of the image and the noise type is non-Gaussian, PCA can only recover the first two or three components. $e$PCA improves the estimation and is able to extract the top 7 eigenimages. Moreover, steerable PCA and steerable $e$PCA achieve much better estimation of the underlying true eigenimages for a given sample size. Steerable $e$PCA achieves the best performance in estimating both the eigenvalues and eigenimages.

\begin{figure*}
\captionsetup[subfloat]{farskip=2pt,captionskip=1pt}
	\begin{center}
		\subfloat[Single Image NLPCA,  MSE$= 0.0013$ ]{
			\includegraphics[width = 0.3\textwidth]{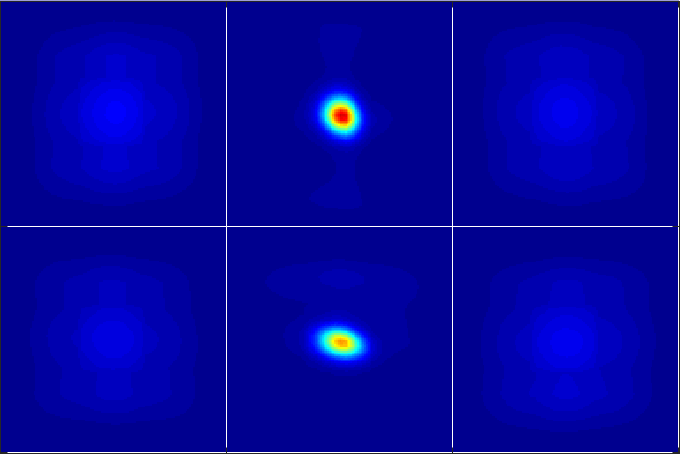}
			\label{fig:denoise_NLPCA}
		}
		\subfloat[PCA, $n = 1000$, MSE$= 1.29 \times 10^{-4}$]{
			\includegraphics[width = 0.3 \textwidth]{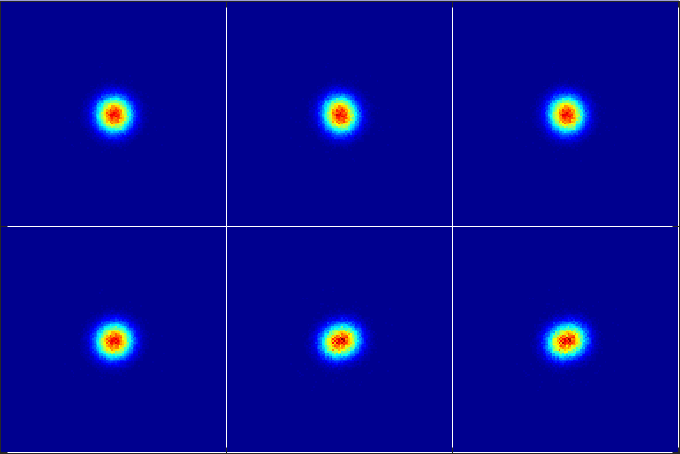}
			\label{fig:denoise_pca_n1000}
		}
		\subfloat[sPCA, $n = 1000$, MSE$= 5.43 \times 10^{-5}$]{
			\includegraphics[width = 0.3 \textwidth]{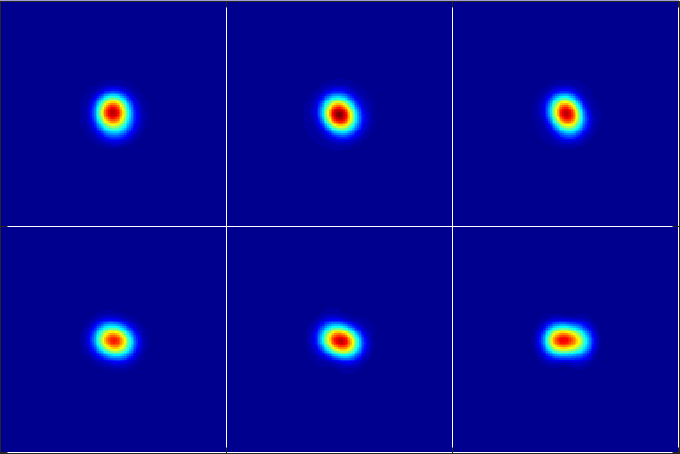}
			\label{fig:denoise_spca_1000}
		} \\
		\subfloat[$e$PCA, $n = 1000$, MSE$= 1.38 \times 10^{-4}$]{
			\includegraphics[width = 0.3\textwidth]{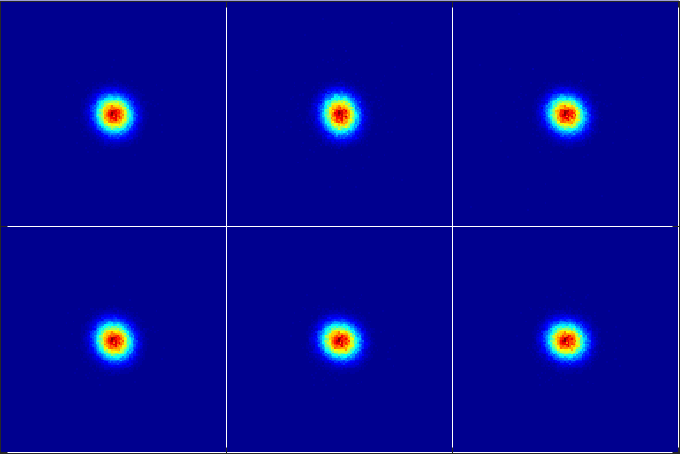}
			\label{fig:denoise_epca_1000}
		}
		\subfloat[s$e$PCA, $n = 1000$, MSE$=2.98\times 10^{-5}$]{
			\includegraphics[width = 0.3\textwidth]{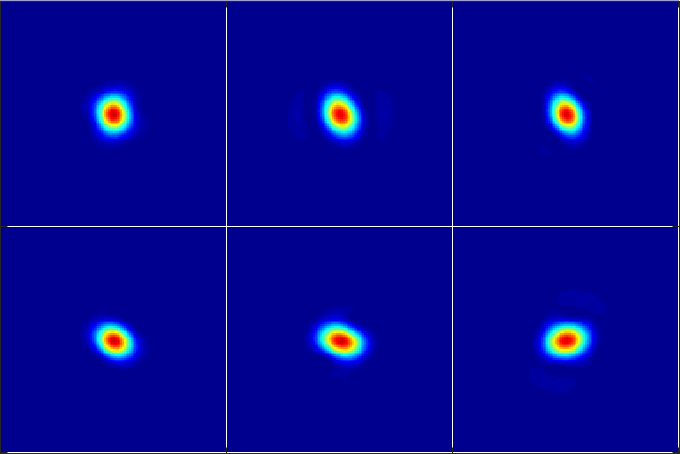}
			\label{fig:denoise_sepca_1000}
		}
		\subfloat[PCA, $n = 70000$, MSE$= 1.85 \times 10^{-4}$]{
			\includegraphics[width = 0.3\textwidth]{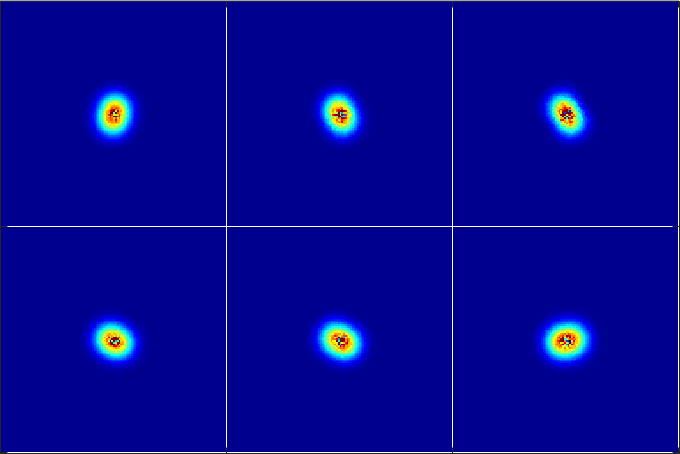}
			\label{fig:denoise_pca_n70000}
		} \\
		\subfloat[sPCA, $n = 70000$, MSE$= 5.19 \times 10^{-5}$]{
			\includegraphics[width = 0.3\textwidth]{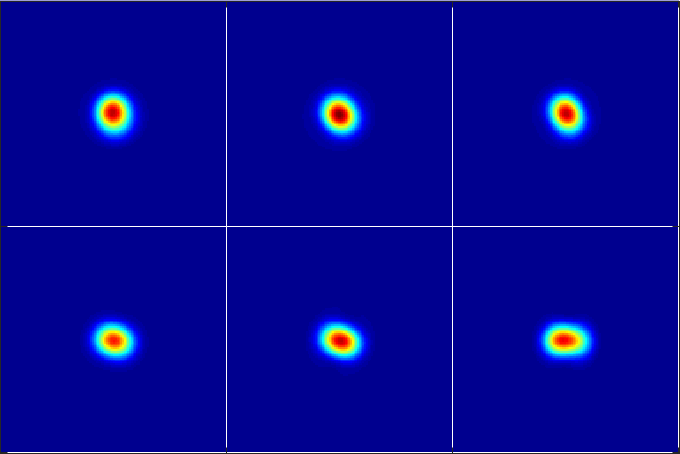}
			\label{fig:denoise_spca_n70000}
		}
		\subfloat[$e$PCA, $n = 70000$, MSE$= 4.16 \times 10^{-5}$]{
			\includegraphics[width = 0.3\textwidth]{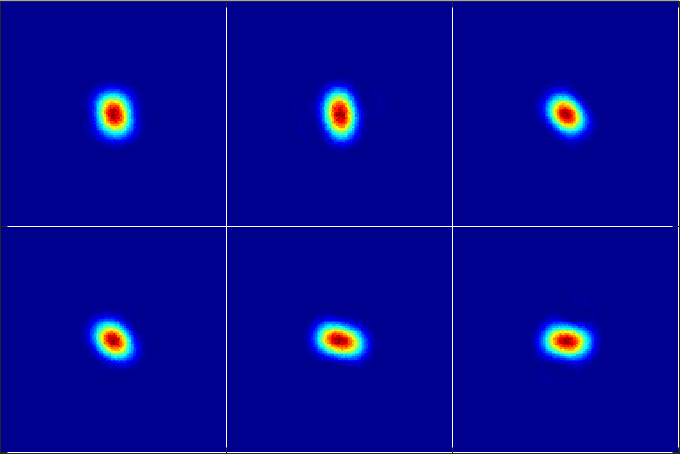}
			\label{fig:denoise_epca_70000}
		} 
		\subfloat[s$e$PCA, $n = 70000$, MSE$=2.97\times 10^{-5}$]{
			\includegraphics[width = 0.3\textwidth]{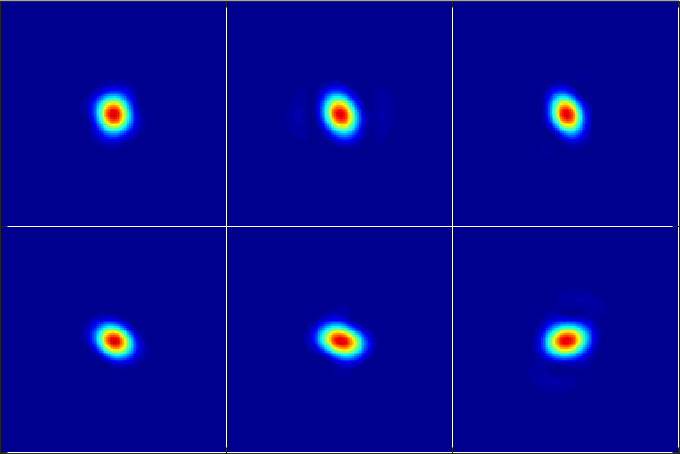}
			\label{fig:denoise_sepca_70000}
		} 
	\end{center}
	\caption{Sample denoised images of the XFEL dataset illustrated in Fig.~\ref{fig:data} using NLPCA, PCA, sPCA, $e$PCA, s$e$PCA. Image size is $128 \times 128$.}
	\label{fig:denoising}
\end{figure*} 

\begin{figure}
	\begin{center}
		\includegraphics[width = 0.8\columnwidth]{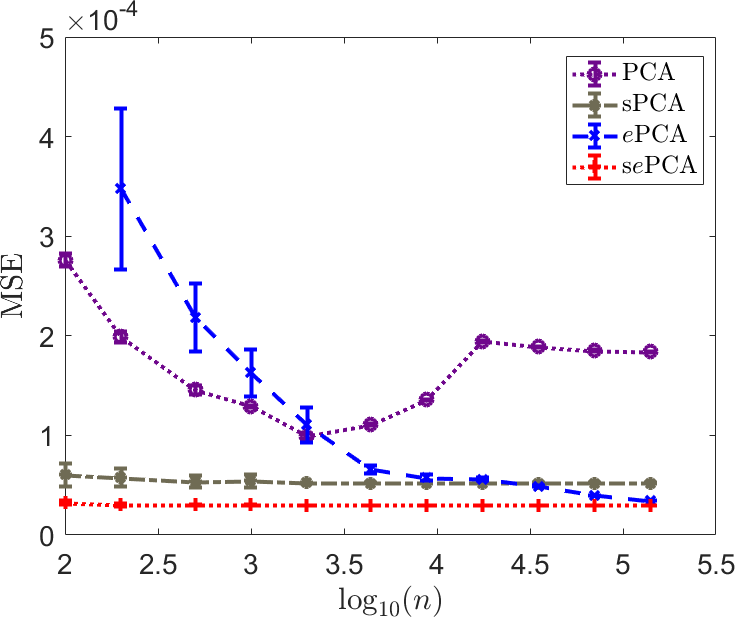}
	\end{center}
	\caption{Comparing the denoising quality of the XFEL dataset ($p = 16384$) with various number of images.}
	\label{fig:denoising_mse}
	\vspace{-0.5cm}
\end{figure} 

Furthermore, we compare the operator norms and Frobenius norms of the difference between the covariance estimates and the true covariance matrix. Fig.~\ref{fig:xfel_err_cov_est} shows that steerable $e$PCA significantly improves the covariance estimation, especially when the sample size is small.

\begin{figure*}
\captionsetup[subfloat]{farskip=2pt,captionskip=1pt}
	\begin{center}
		\subfloat[ Single Image NLPCA, MSE$= 7.43$ ]{
			\includegraphics[width = 0.27\textwidth]{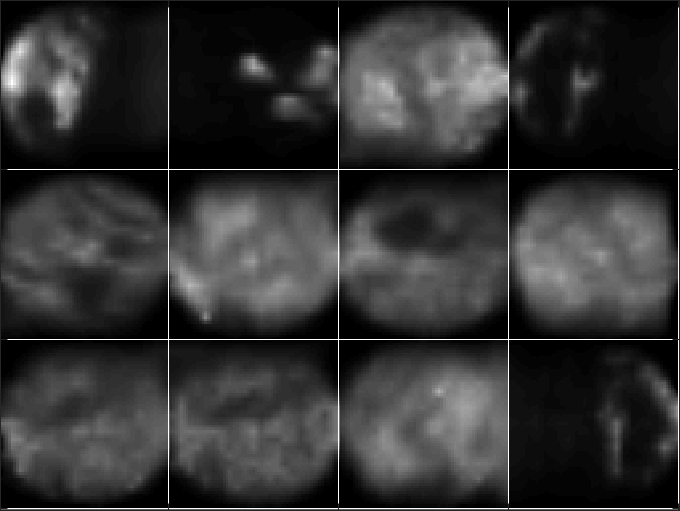}
			\label{fig:face_NLPCA}
		}
	   \subfloat[ PCA, $n = 2^{10}$, MSE$= 0.523$ ]{
			\includegraphics[width = 0.27\textwidth]{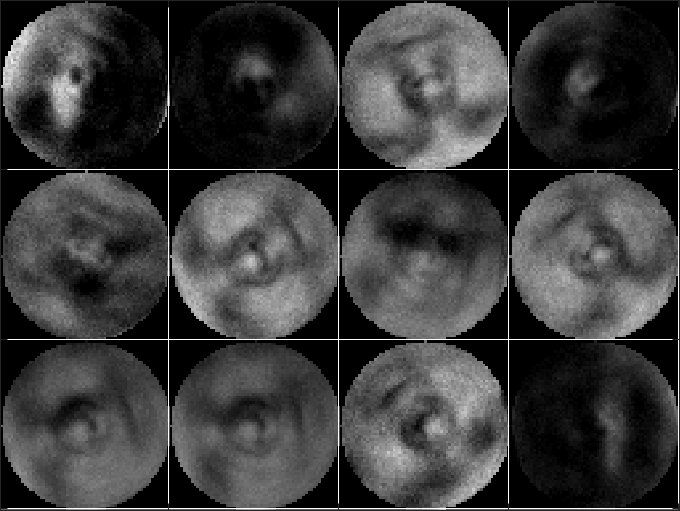}
			\label{fig:face_pca_small}
		} 
	   \subfloat[ sPCA, $n = 2^{10}$,  MSE$= 0.311$ ]{
			\includegraphics[width = 0.27\textwidth]{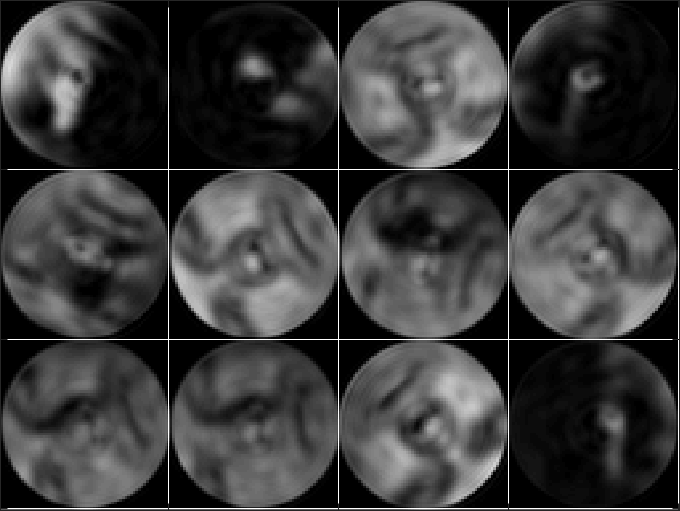}
			\label{fig:face_spca_small}
		} \\
		\subfloat[$e$PCA, $n = 2^{10}$, MSE$= 0.509$]{
			\includegraphics[width = 0.27\textwidth]{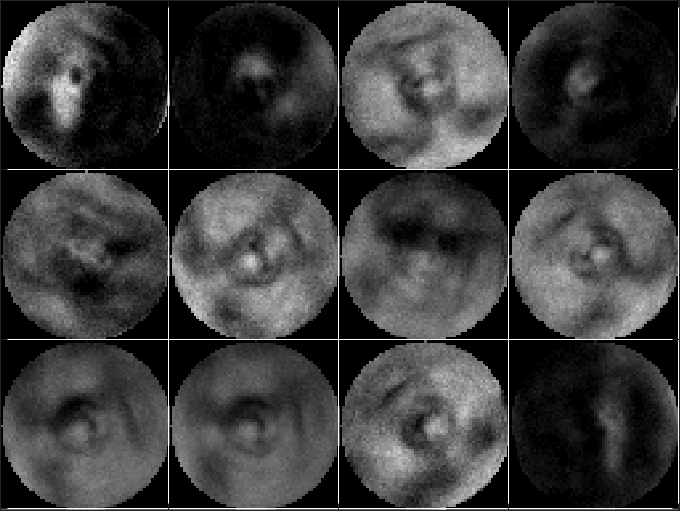}
			\label{fig:face_epca_small}
		}
		\subfloat[ s$e$PCA, $n = 2^{10}$, MSE$= 0.237$ ]{
			\includegraphics[width = 0.27\textwidth]{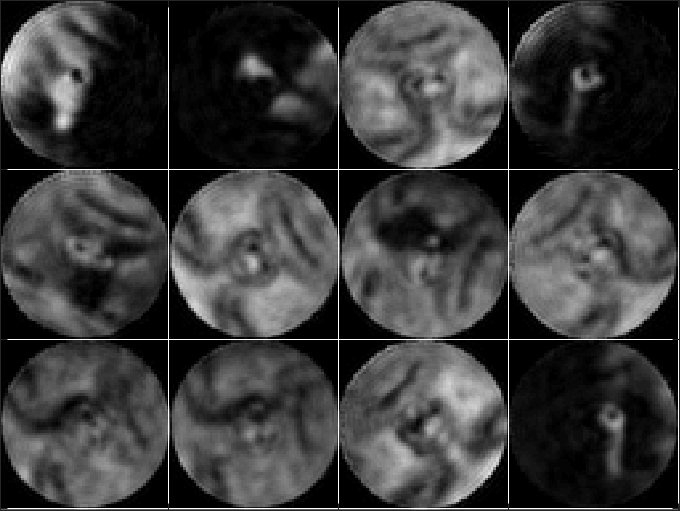}
			\label{fig:face_sepca_small}
		} 
        \subfloat[ PCA, $n = 2^{15}$, MSE$= 0.255 $]{
			\includegraphics[width = 0.27\textwidth]{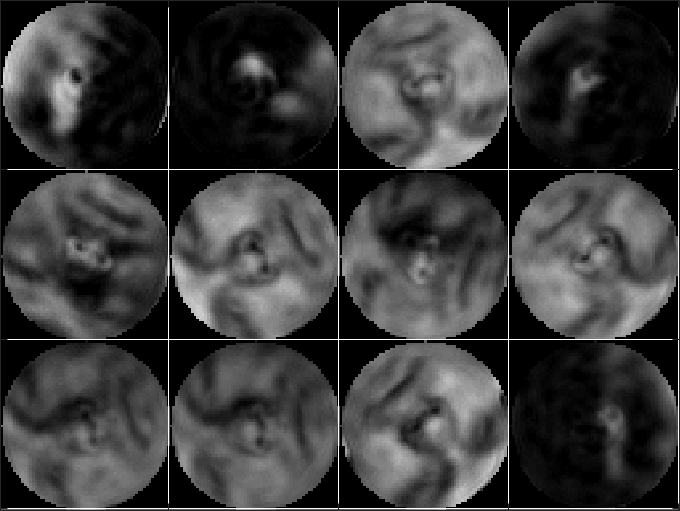}
			\label{fig:face_pca_large}
		} \\
	   \subfloat[ sPCA, $n = 2^{15}$, MSE$= 0.307 $]{
			\includegraphics[width = 0.27\textwidth]{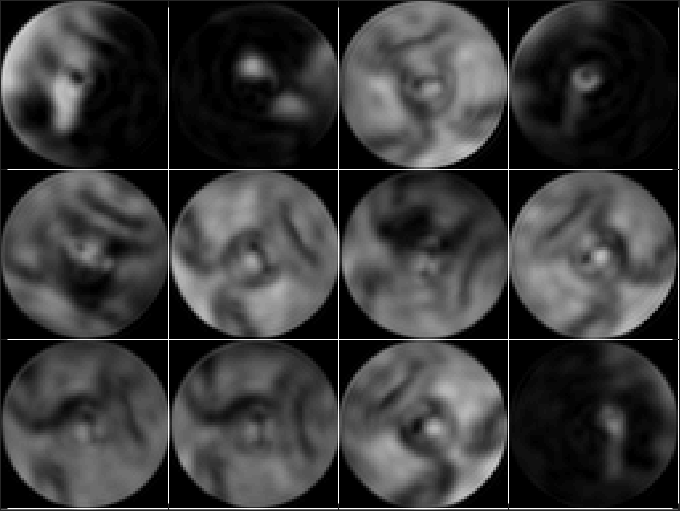}
			\label{fig:face_spca_large}
		} 
		\subfloat[ $e$PCA, $n = 2^{15}$,  MSE$= 0.239 $ ]{
			\includegraphics[width = 0.27\textwidth]{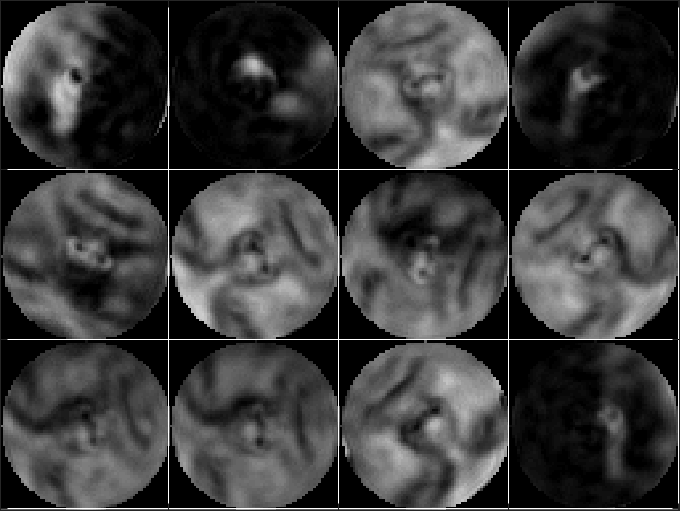}
			\label{fig:face_epca_large}
		} 
		\subfloat[ s$e$PCA, $n = 2^{15}$, MSE$= 0.223 $ ]{
			\includegraphics[width = 0.27\textwidth]{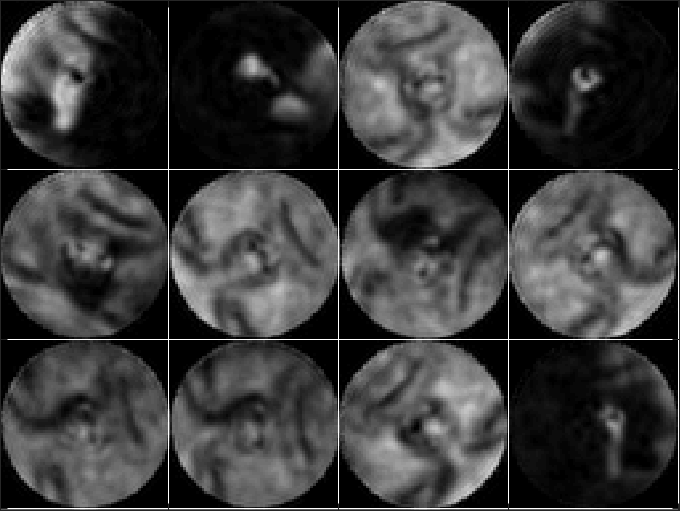}
			\label{fig:face_sepca_large}
		} 
	\end{center}
	\caption{Sample denoised images of the face dataset illustrated in Fig.~\ref{fig:face_data} using NLPCA, PCA, sPCA, $e$PCA, and s$e$PCA. Image size is $64 \times 64$. The intensity range is [0, 2.6] for \protect \subref{fig:face_NLPCA}, and is [0, 11.1] for \protect \subref{fig:face_pca_small}--\protect \subref{fig:face_sepca_large}. }
	\label{fig:face_denoising}
	\vspace{-0.6cm}
\end{figure*} 

For experiments using $e$PCA, we use a permutation bootstrap-based method to estimate the rank of the covariance matrix, following e.g. \cite{Landgrebe2002}. By randomly permuting each column of the mean-subtracted data matrix, we completely destroy structural information including linear structure, while the noise statistics remain the same (see \cite{dobriban17factor,Dobriban2018} for an analysis). Singular values of the randomly permuted matrices reveal what should be the largest covariance matrix eigenvalues that correspond to noise, up to a user-selected confidence level $\rho$. This can replace the other rank estimation methods that assume Gaussian distribution when the noise model is non-Gaussian, such as in our case. Empirically, we observe that $\rho = 0.1$ gives the best performance in covariance estimation. For steerable $e$PCA, we estimate the number of components using the right edge of the Mar{\v c}enko-Pastur distribution for homogenized covariance matrices $S^{(k)}_{h}$ and include only the components whose scaling factor $\hat{\alpha}^{(k)}$ are above zero.

Steerable $e$PCA is able to recover more signal principal components from noisy images than PCA, steerable PCA, and $e$PCA (see Fig.~\ref{fig:estimated_ranks}). When the sample size $n = 1000$, the mean number of estimated components is $11$ and $49$ for $e$PCA  and steerable $e$PCA respectively. For $n = 70000$, the estimated number of components is $59$ and $81$ for $e$PCA and steerable $e$PCA respectively.  Fig.~\ref{fig:runtime} shows that steerable $e$PCA is more efficient than $e$PCA and PCA. Because steerable $e$PCA contains extra steps such as prewhitening, recoloring, and scaling, its runtime is slightly longer than steerable PCA. When $n = 140000$, steerable $e$PCA is 8 times faster than $e$PCA. 

In addition to the XFEL diffraction intensity data, we include a natural image dataset--Yale B face database ~\cite{lee2005acquiring,georghiades2001few}--to illustrate the efficacy of the proposed method. The database contains 5760 single light source images of 10 subjects with 9 poses. 
For every subject in a particular pose, $64$ images were captured under different ambient illuminations. The original images of a single face under different lighting conditions inhabit an approximately 9-dimensional linear space~\cite{Basri2003}. In the experiments, we take one subject at a particular pose with 64 different illumination conditions. We downsample the original images to the size of $64 \times 64$ pixels and scale the intensity such that the mean photon count for the whole dataset is $2.3$ photons per pixel (i.e., ``faces in the dark''). Since we need to rotate the images, we mask images by a disk of radius $R = 32$. We uniformly sample $n$ clean images from the original images and apply a random in-plane rotation to each sample. Fig.~\ref{fig:face_data} shows 12 samples of the clean data and the corresponding noisy observations. The corresponding principal components are illustrated in Fig.~\ref{fig:eigVec_face}. The true eigenimages are evaluated using the clean images rotated at every 0.7 degrees.

\subsection{Denoising}
We compare the denoising effects of steerable $e$PCA with PCA, steerable PCA, $e$PCA, and patch-based single image non-local PCA (NLPCA)~\cite{salmon2014poisson}, by the mean squared error, MSE$:= (pn)^{-1} \sum_{i=1}^n \|\hat{X}_i - X_i \|^2 $. 
We perform ``$e$PCA denoising'' using the empirical best linear predictor (EBLP)~\cite{LiuEtAl2017}, which had been shown to outperform ``PCA denoising,'' i.e., orthogonal projection onto sample or $e$PCA / heterogenized eigenimages, as well as the exponential family PCA method proposed by~\cite{Collins2001}.  Note that in our implementation of $e$PCA Wiener-type filtering, to avoid inverting a singular matrix (when some coordinates have 0 sample mean and sample variance), we compute $\diag[\bar Y] + S_s$ with regularization, $\diag[\bar Y ] +S_s \leftarrow S_s +(1-\epsilon)\diag[\bar Y ]+\epsilon m I$ where $\epsilon = 0.1$, $m = \frac{1}{p} \bar Y \cdot \vect{1}$, and $I$ is the $p \times p$ identity matrix. The number of components are estimated by the permutation rank estimation as described in the previous section.  

Fig.~\ref{fig:denoising} shows some examples of denoised XFEL images for sample size $n = 1000$ and $70000$. For robustness, we repeated the numerical experiment for another dataset simulated from the small protein chignolin (Protein Data Bank entry 1UAO) and obtained qualitatively similar results. Steerable $e$PCA is able to recover the images with lower MSEs compared to PCA, steerable PCA, and $e$PCA (see Fig.~\ref{fig:denoising_mse}), especially when the sample size $n$ is small. 

\begin{figure}
	\begin{center}
		\includegraphics[width = 0.8\columnwidth]{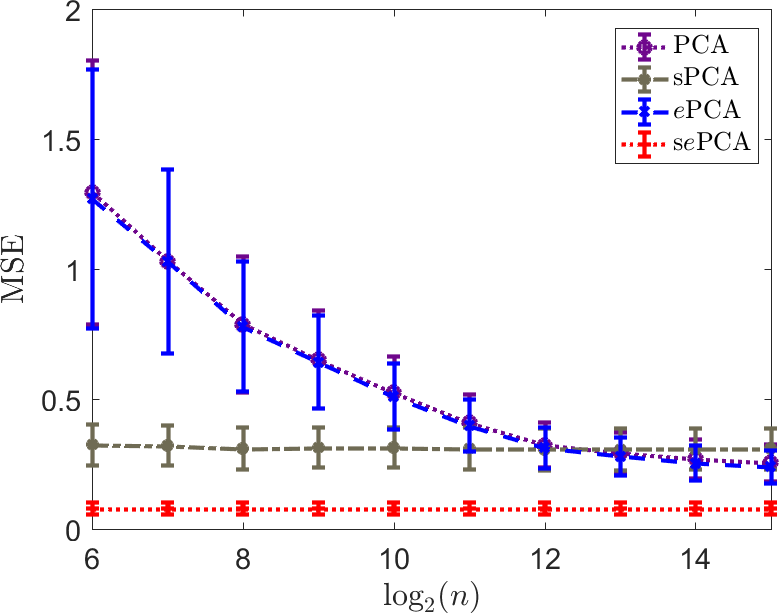}
	\end{center}
	\caption{Comparing the denoising quality of the face dataset ($p = 4096$) with various number of images.}
	\label{fig:face_denoising_mse}
\end{figure} 

\begin{figure}
    \centering
    \subfloat[ XFEL images ]{
    \includegraphics[ width = 0.48 \columnwidth ]{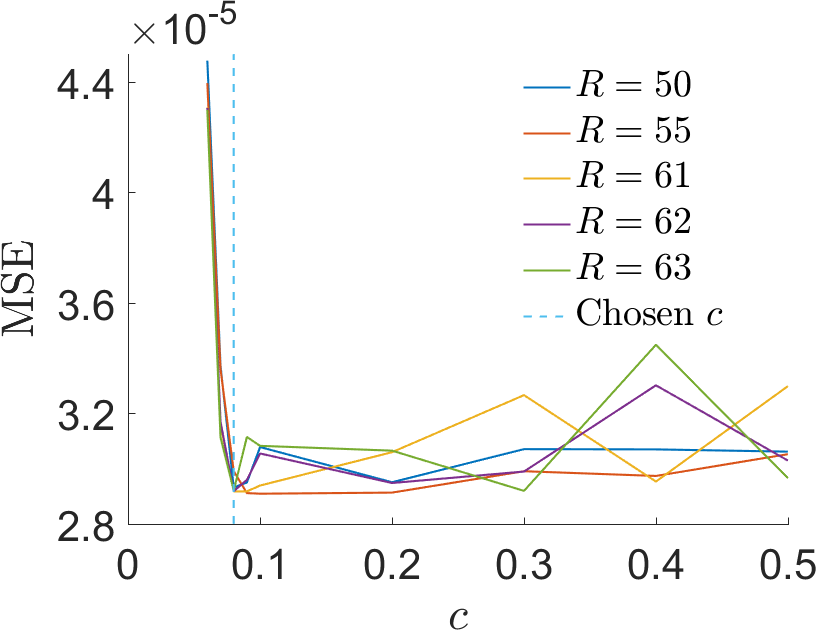}
    \label{fig:xfel_test_c}
    }
    \subfloat[ Rotated Yale B face ]{
     \includegraphics[ width = 0.48 \columnwidth ]{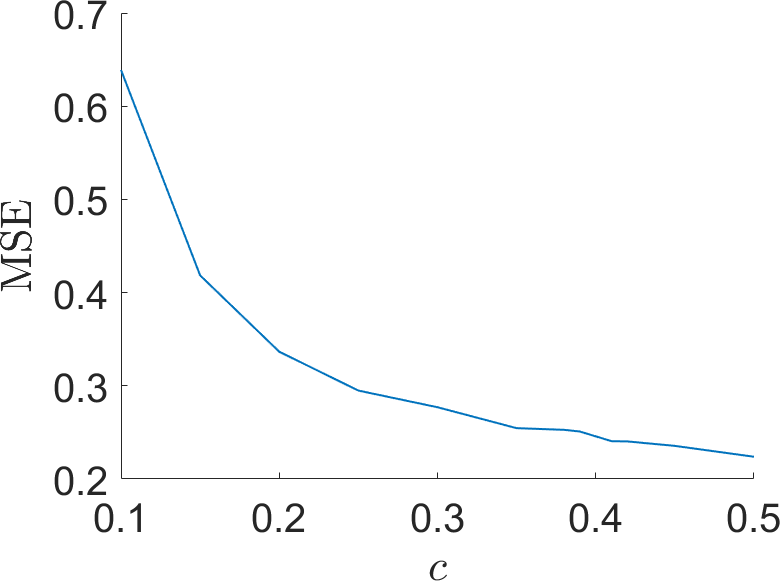}
     \label{fig:face_test_c}
    }
    \caption{Sensitivity of the denoising performance over the parameters $c$ and $R$. For the face data, we cropped the images by the disc of radius $R = 32$. }
    \label{fig:test_c_R}
\end{figure}

Fig.~\ref{fig:face_denoising} shows examples of the denoised face images for sample size $n = 2^{10}$ and $2^{15}$. We use the permutation rank estimation described in the previous section with confidence level $\rho = 10$ for PCA, sPCA, and $e$PCA. In this example, we see that when the number of images is small, PCA and $e$PCA can only recover rough contours of the faces, whereas sPCA and s$e$PCA are able to recover finer details in the faces. The MSEs of the denoised images with varying number of samples are plotted in Fig.~\ref{fig:face_denoising_mse}. In this experiment, the data only contains 64 unrotated clean images and the mean photon count is 2.3 per pixel. Compared to the diffraction patterns with mean photon count 0.01 per pixel. Therefore, s$e$PCA is able to sufficiently capture the clean data subspace with only a small number of images, resulting in a much flatter MSE curve with respect to the number of images.

We also test how sensitive our method is to the choice of parameters, i.e. the band limit $c$ and the support radius $R$. For the diffraction intensity data, we use $n = 10000$ diffraction patterns with mean photon count 0.01 per pixel. The underlying clean diffraction patterns are very smooth. We vary the band limit from $c = 0.06$ to $0.5$ and the support radius $R$ ranges from 50 to 63. With the chosen parameters indicated in Fig.~\ref{fig:cR}, the MSE for the denoised image is small (see Fig.~\ref{fig:xfel_test_c}). Although the MSEs vary when we increase the parameter $c$ above the chosen band limit and change $R$ within the range of 50 to 63, the errors are still relatively small ($2.9-3.5 \times 10^{-5}$), compared to above $5 \times 10^{-5}$ with $e$PCA and sPCA. For the face data, we compare the results with $n = 8192$ noisy images with 2.3 photons per pixel. Since we crop out the region within a disk of radius $R = L/2$, the parameter $R$ is fixed to be 32 and we vary the band limit $c$ from $0.1$ to $0.5$. We observe that choosing $c = 0.5$ provides the best denoising performance (see Fig.~\ref{fig:face_test_c}), and this agrees with the fact that the face images contain more high-frequency information than the XFEL data.

\section{Conclusion}

We presented steerable $e$PCA, a method for principal component analysis of a set of low-photon count images and their uniform in-plane rotations.  
This work has been mostly motivated by its application to XFEL, but is relevant to other low-photon count imaging applications. 
The computational complexity of the new algorithm is $O(nL^3 + L^4)$, whereas that of $e$PCA is $O(\min(nL^4 + L^6, n^2 L^2 + n^3))$. Incorporating rotational invariance allows more robust estimation of the true eigenvalues and eigenvectors. Our numerical experiments showed that steerable $e$PCA more accurately estimates the covariance matrix and achieves better denoising results. Finally, we remark that the Fourier-Bessel basis can be replaced with other suitable bases, for example, the 2-D prolate spheroidal wave functions (PSWF) on a disk~\cite{Slepian, landa2017steerable}. Nevertheless, our method has certain limitations. For example, if the noise distribution is not specified in advance, it is hard to estimate the noise covariance $\E \diag[A''(\omega)]$, which will affect the homogenization step of the proposed method. In addition, the optimal shrinkage function for the covariance estimation depends on how we measure the error, for example using Frobenius norm, operator norm, or other criteria.  The corresponding theoretical guarantee in the non-Gaussian case is still an open problem. Thus, the statistical optimality is beyond the scope of this paper. For future work, we will study how sensitive the proposed method is to the misspecification of the noise distribution and the statistical optimality of our procedure in the non-Gaussian case. 

\section*{Acknowledgments}
The authors would like to thank associate editor and the anonymous referees for their valuable comments. The authors would also like to thank Edgar Dobriban and Boris Landa for discussions. ZZ was partially supported by National Center for Supercomputing Applications Faculty Fellowship and University of Illinois at Urbana-Champaign College of Engineering Strategic Research Initiative. AS was partially supported by Award Number R01GM090200 from the NIGMS, FA9550-17-1-0291 from AFOSR, Simons Investigator Award, the Moore Foundation Data-Driven Discovery Investigator Award, and NSF BIGDATA Award IIS-1837992.

\bibliographystyle{IEEEtran}
\bibliography{FBsPCA,sePCA,ePCA_ref}

\end{document}